\documentclass[10pt,twocolumn,letterpaper]{article}

\usepackage[pagenumbers]{cvpr} 

\usepackage{graphicx}
\usepackage{amsmath}
\usepackage{amssymb}
\usepackage{booktabs}
\usepackage{xcolor}

\usepackage[pagebackref,breaklinks,colorlinks]{hyperref}

\usepackage[capitalize]{cleveref}
\crefname{section}{Sec.}{Secs.}
\Crefname{section}{Section}{Sections}
\Crefname{table}{Table}{Tables}
\crefname{table}{Tab.}{Tabs.}

\def\cvprPaperID{8319} 
\def\confName{CVPR}
\def\confYear{2022}

\newif\ifdraft
\draftfalse
\drafttrue

\ifdraft
\newcommand{\PF}[1]{{\color{red}{\bf PF: #1}}}

\newcommand{\ND}[1]{{\color{blue}{\bf ND: #1}}}

\else
\newcommand{\PF}[1]{}

\newcommand{\AL}[1]{}

\newcommand{\ND}[1]{}

\fi

\newcommand{\pale}[0]{\textit{{PartAL}}}
\newcommand{\parag}[1]{\vspace{-3mm}\paragraph{#1}}

\newcommand{\bL}{\textbf{L}}
\newcommand{\bU}{\textbf{U}}

\newcommand{\bY}{\mathbf{Y}}

\newcommand{\bx}{\mathbf{x}}
\newcommand{\by}{\mathbf{y}}
\newcommand{\bz}{\mathbf{z}}

\begin{document}

\title{PartAL: Efficient Partial Active \\ Learning in Multi-Task Visual Settings}

\author{
    Nikita Durasov\textsuperscript{1}$^*$
    \quad
    Nik Dorndorf\textsuperscript{2}$^*$
    \quad
    Pascal Fua\textsuperscript{1}\\
    \textsuperscript{1}Computer Vision Laboratory, EPFL, {\tt\small \{name.surname\}@epfl.ch}\\
    \textsuperscript{2}RWTH Aachen, {\tt\small \{name.surname\}@rwth-aachen.de}\\
}

\maketitle
\def\thefootnote{*}\footnotetext{These authors contributed equally to this work}\def\thefootnote{\arabic{footnote}}


\begin{abstract}

Multi-task learning is central to many real-world applications. Unfortunately, obtaining labelled data for all tasks is time-consuming, challenging, and expensive. Active Learning (AL) can be used to reduce this burden. Existing techniques typically involve picking images to be annotated and providing annotations for all tasks. 

In this paper, we show that it is more effective to select not only the images to be annotated but also a subset of tasks for which to provide annotations at each AL iteration. Furthermore, the annotations that are provided can be used to guess pseudo-labels for the tasks that remain unannotated. We demonstrate the effectiveness of our approach on several popular multi-task datasets.

\end{abstract}


\section{Introduction}

\textit{Multi Task Learning} (MTL)  seeks to train deep networks that can solve several  tasks simultaneously given a single input~\cite{Ruder17,Vandenhende21}. Each task usually corresponds to a different modality, such as when estimating depth, normal, and saliency maps from a single RGB image. Not only is performing MTL faster than handling each task separately, but it also tends to deliver higher performance by virtue of sharing information across modalities. One major roadblock, however, is that training a single-task network in a fully supervised manner often requires large amounts of data and training a multi-task one necessitates even more. 


\begin{figure}[h!]
    \includegraphics[width=0.5\textwidth]{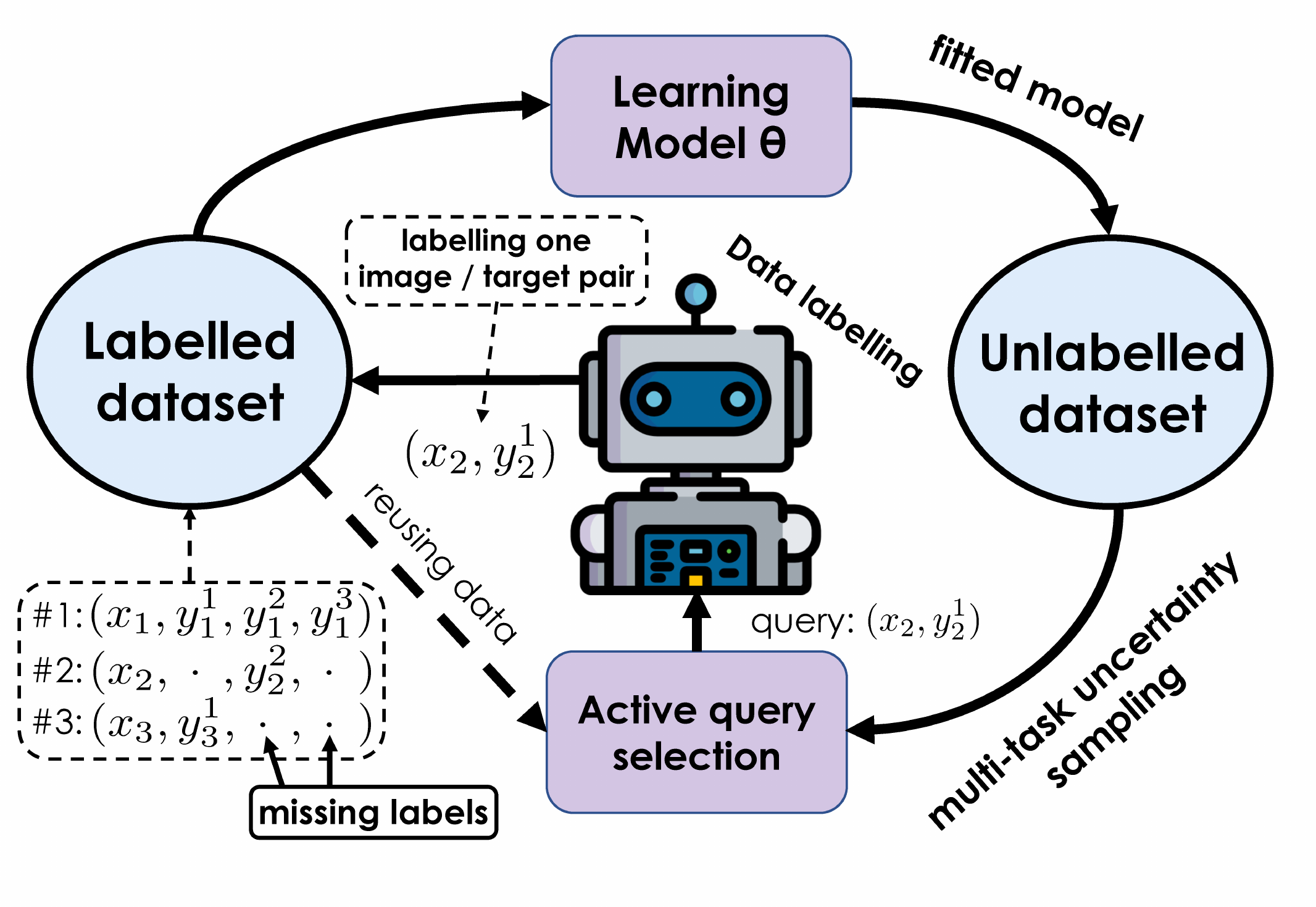}
    \centering
    \vspace{-9mm}
    \caption{
    \textbf{\pale{}'s active learning scheme.} Instead of selecting images to be annotated and providing labels for {\it all} modalities, we provide labels only for the {\it most relevant} ones. In the end, this uses labels more effectively and significantly reduces the total number that needs to be provided.}
    \vspace{-2mm}
    \label{fig:mtask}
\end{figure}

In machine learning field, \textit{Active Learning} (AL)~\cite{Settles09a} has been widely used to reduce the amount of annotated data required to train single-task networks but {\it not} multi-task ones. The few techniques that use AL for multi-tasking purposes~\cite{Reichart08,Qi08} do not naturally generalize to deep-learning.

\begin{figure}[h!]
    \includegraphics[width=0.48\textwidth]{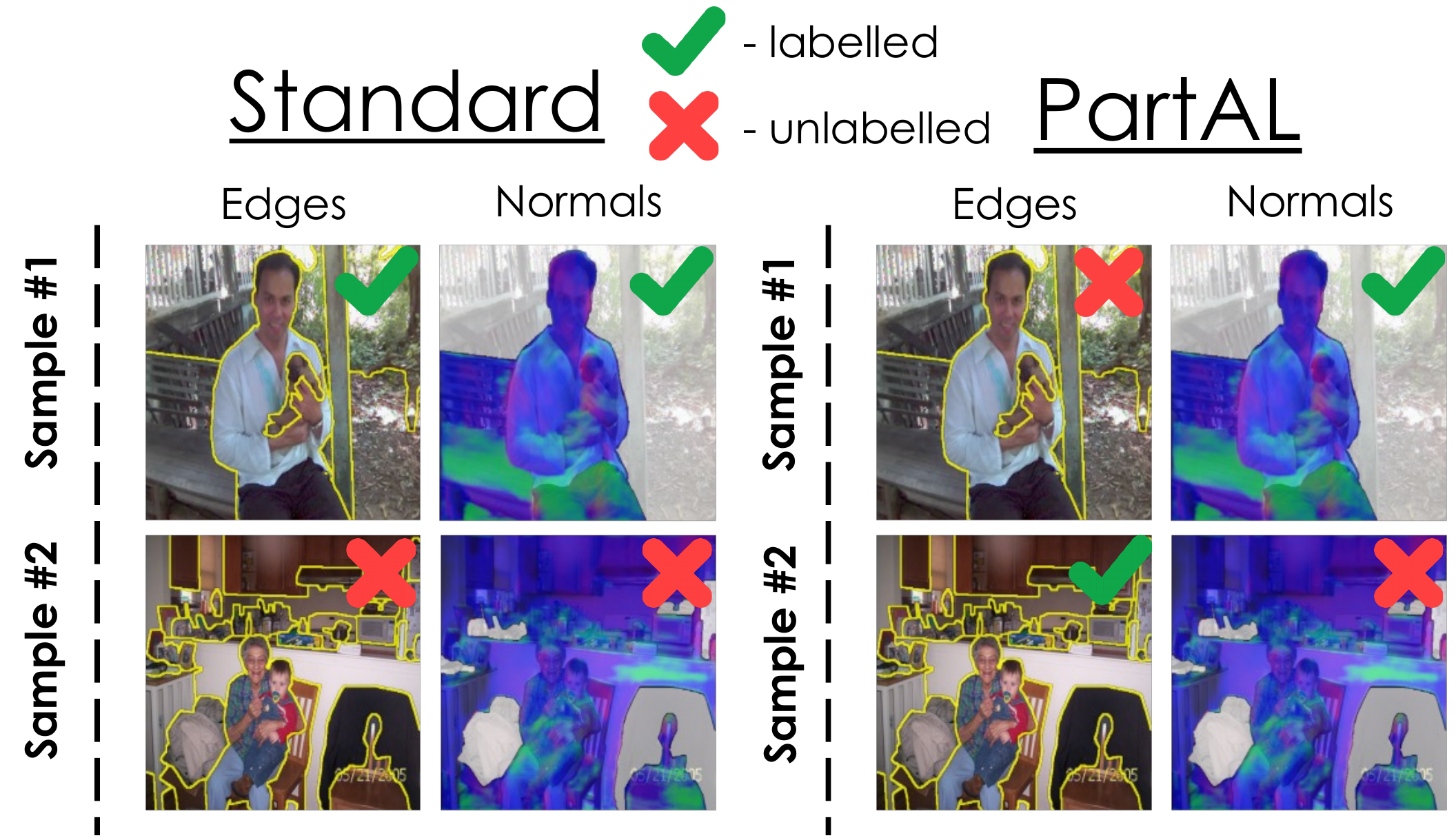}
    \centering
    \vspace{-7mm}
    \caption{\small
    \textbf{\pale{}'s vs standard active learning labelling pipelines.} Unlike conventional AL pipelines, \pale{} allows the user to provide partial labels, that is, ground-truth for some modalities and not others. In this particular example, standard method decides to provide all modalities labels for the first sample and no labels for the second. By contrast, our method has the flexibility to provide only one modality per sample.
    }
    \label{fig:al}
\end{figure}

An obvious way to incorporate standard AL techniques into MTL would be to find images about which the network is most uncertain and, for these, supply image annotations for {\it all} tasks the network is supposed to perform. However, this would be inefficient because it does not exploit the specificities of MTL. We show that we can achieve better performance and a lower annotation cost by, at each AL iteration, supplying annotations only for some of the tasks for any image under consideration. In effect, our algorithm determines {\it both} which images and which modalities should be labelled for these images, as depicted by Fig.~\ref{fig:al}. We will refer to the annotations we provide as  \textit{partial labels}, as opposed to \textit{full labels} that would involve providing annotations for all modalities. To further boost performance, we reuse labelled modalities in further active learning steps to estimate the unlabelled ones with higher accuracy. We will refer to this approach as \pale{} and we will show that it produces better trained networks while requiring fewer annotations than other approaches to active learning.


\section{Related Work}
\label{sec:related}

\textit{Active Learning} (AL) is one of the most efficient and popular approaches to reducing the amount of annotations required to train deep networks. However, it has been deployed mostly for the purpose of training single-task networks as opposed to multi-task ones. In what follows, we provide background on active and multi-task learning.

\subsection{Multi-Task Learning}

Modern neural networks excel at tackling many different computer vision tasks, such as image segmentation and depth estimation. The idea underpinning multi-task learning is that if one can train a network to perform these tasks jointly, as opposed to individually, performance will increase because the various tasks are correlated and intermediate features that are good for one ought to be good for the others. For example, many depth boundaries are also segmentation boundaries and finding one helps finding the other. Thus, forcing the features to be similar for all tasks tends to yield better generalization properties. 

Multi-task deep learning models are often considered as belonging to one of two classes~\cite{Ruder17}. The first includes networks that use an encoder to produce a common representation that is then decoded by separate decoders, one per task~\cite{Caruana97,Kokkinos17,Kendall18}. The second category comprises separate network branches for each task, each with their own parameters that are constrained not to vary too much across tasks or to share information with each other via cross-talk~\cite{Duong15,Yang16d,Misra16a,Gao19c}.

\begin{figure}[h!]
    \centering
    \includegraphics[width=0.47\textwidth]{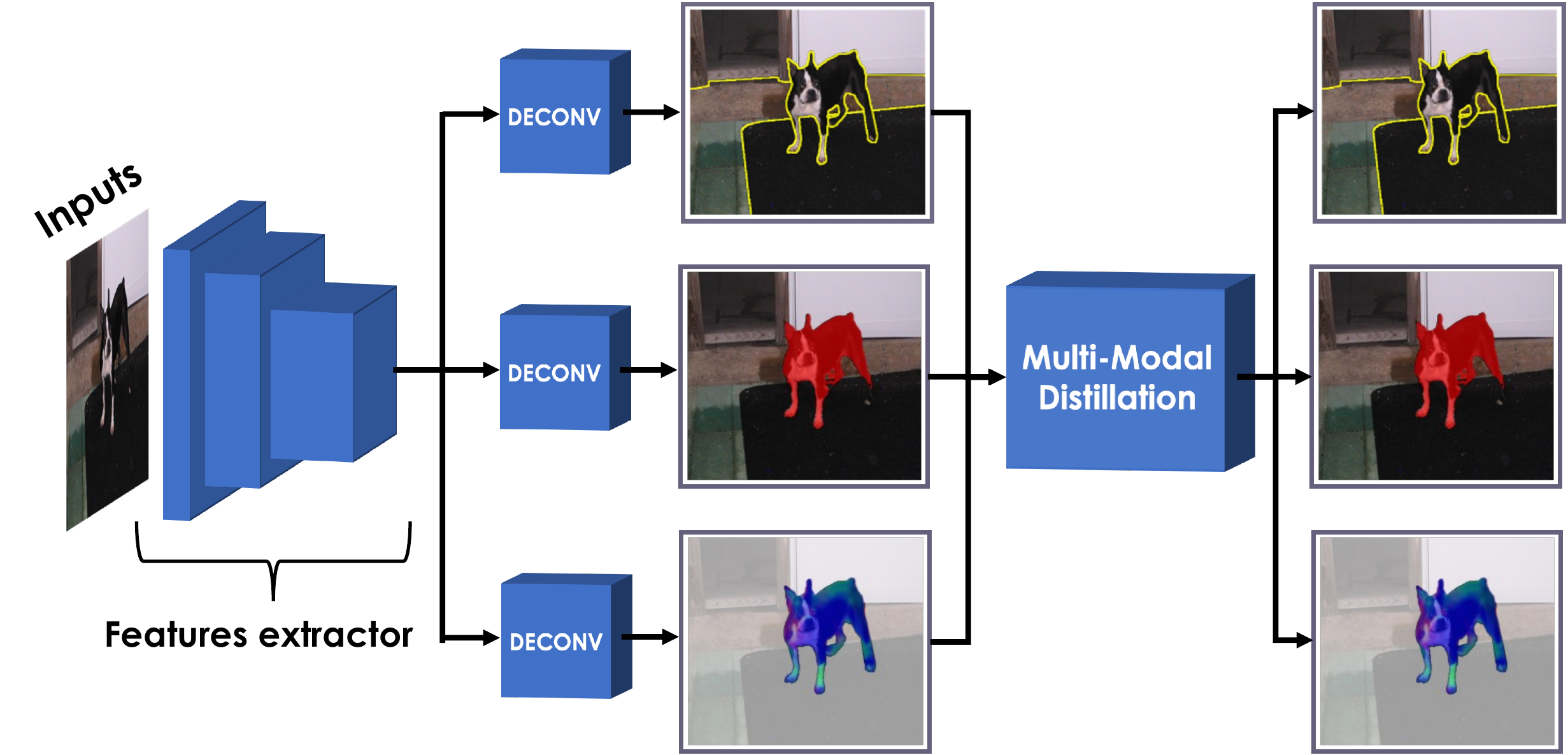}
    \caption{\small {\bf PAD-Net for multi-task inference.} Given a single RGB image as an input, PAD-Net produces a set of intermediate predictions that are fed to a distillation module. The output is a final set of predictions that pool information from all modalities. At inference time, we modify the network so it is possible to replace some of the intermediate predictions by ground-truth images if they happen to be available.}
    \label{fig:padnetwork}
\end{figure}

There are also approaches that combine elements from both categories~\cite{Xu18c,Zhang19f,Vandenhende20}. This can involve sharing information about tasks both in the shared backbone and in the decoder. The PAD-Net~\cite{Xu18c} architecture depicted in Fig.~\ref{fig:padnetwork} is a good example of this and one of the most widely used multi-task architectures~\cite{Vandenhende21}.  PAD-Net stands for Prediction-and-Distillation Network and has a two-stage architecture. Its first stage takes an input RGB image and generates an intermediate representation that is passed to dedicated decoders to generate predictions for the individual tasks. The second stage then performs multi-modal distillation on these predictions and outputs refined versions of the predictions that should be consistent with each other. This merges information from various sources to enhance the final predictions.

\subsection{Single-Task Active Learning}

Active Learning (AL) makes it possible to streamline the data labelling process by only annotating the most informative samples~\cite{Settles09a}. It has proved useful in computer vision~\cite{Li13f,Gal17,Beluch18,Kaushal19}, natural language processing~\cite{Settles08b,Siddhant18,Dor20}, and many other engineering domains~\cite{Hao20b,Flores20,Durasov21b}. 

Given a trained predictor, the most common AL approach to it relies on uncertainty sampling~\cite{Lewis94}, that is, asking an annotator to provide information about the instances about which the predictor is most uncertain. This requires the predictor to estimate its own uncertainty. When the predictor is a neural network,  MC-Dropout~\cite{Srivastava14,Gal16a} and Deep Ensembles~\cite{Lakshminarayanan17} have emerged as two of the most popular methods; with Bayesian networks~\cite{MacKay95} being a third alternative.  MC-Dropout involves randomly zeroing out network weights and assessing the effect, whereas Ensembles involve training multiple networks, starting from different initial conditions. In practice, the latter tends to perform better but also to be far more computationally demanding because the training procedure has to be restarted from scratch several times. The method of~\cite{Durasov21a} attempts to get the best of both worlds by using multiple binary masks to zero-out selected network weights, which can be done in a more controlled fashion than in MC-Dropout. 

Other powerful approaches to Active Learning include \textit{Coreset}~\cite{Sener17} and \textit{LearningLoss}~\cite{Yoo19}, which have both proved to be among the  best performing approaches. Coreset tries to find a \textit{core-set}~\cite{Tsang05} of unlabelled samples by choosing a subset of points such that the performance of a model learned over the selected subset is the closest to the performance of the model trained on the whole data. LearningLoss learns to predict what samples should be queried.  Both approaches often outperform the uncertainty-based ones. However, these methods were not designed with multiple tasks in mind, whereas ours is. In our experiments, we quantitatively demonstrate that \pale{} outperforms them. 


\begin{figure}[h!]
    \includegraphics[width=0.5\textwidth]{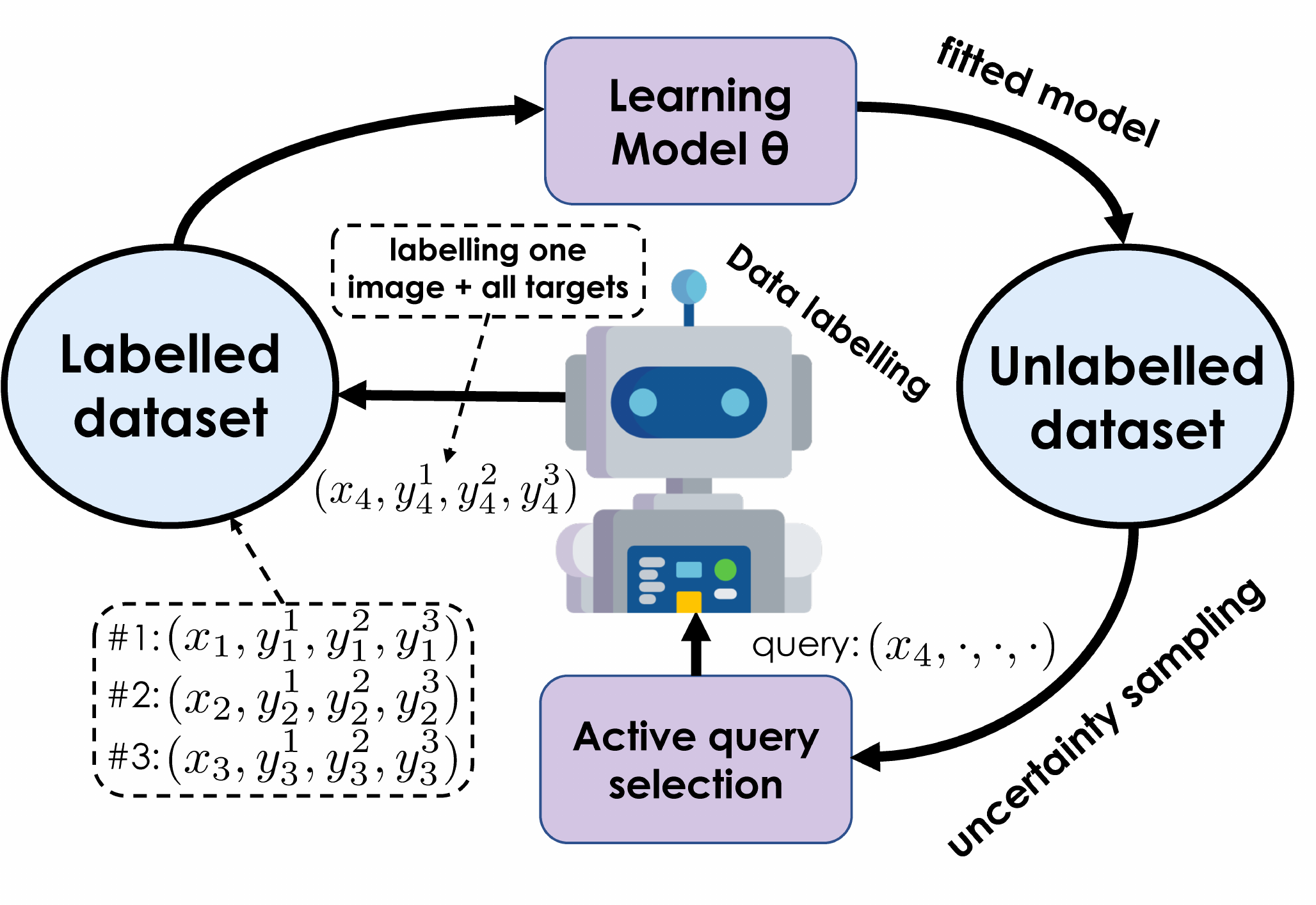}
    \centering
    \vspace{-7mm}
    \caption{
    \textbf{Vanilla active learning scheme.} Following the general single-task setup, existing multi-task methods that rely on AL~\cite{Reichart08,Qi08} perform the standard training and uncertainty sampling steps. While querying new data for labelling, they pick unlabelled images and provide annotations for \textit{all} of the tasks. This is to be contrasted by our approach, as depicted by Fig.~\ref{fig:mtask}.
    }
    \label{fig:vanilla-mtal}
\end{figure}

\subsection{Multi-Task Active Learning}

AL has proven its worth for the purpose of training single-task networks but has never been deployed to train deep multi-task ones. In fact, we know of only two works~\cite{Reichart08,Qi08} that use active learning in a multi-task context and the corresponding schemes are depicted by Fig.~\ref{fig:vanilla-mtal}, to be contrasted with our proposed approach depicted by Fig.~\ref{fig:mtask}. In~\cite{Reichart08}, an active learning pipeline is used to train classifiers to handle two separate problems, entity recognition and syntactic parsing. Even though this method is not fully multi-task because different models are used for each task, it uses active learning in a way that incorporates information about both. Namely, the uncertainty for each unlabelled sample is computed and then used to rank them according to a score that combines both uncertainty estimates. Finally,  only those with the highest score are queried for labelling. This approach ignores the multi-modal nature of the data and labels for all tasks have to be provided for samples that are queried. This is suboptimal because the different targets for one input are correlated and labelling more than one task for any given sample might be overkill. In~\cite{Qi08}, an approach to two-dimensional active learning for multi-class classification is proposed. Instead of labelling all classes for one sample, only single classes are labelled by selecting the most uncertain sample / class pairs. Although effective for classification, there is no obvious way to scale up this approach to more complicated and varied visual tasks, such as depth estimation or segmentation.


\section{Method}
\label{sec:method}

Let us consider a small labelled dataset $\bL = \{(\bx_{i}, \bY_i)\}_{i=1}^{N_L}$ and a much larger unlabelled one $\bU = \{\bx_{i}\}_{i=1}^{N_U}$, where $\bx_i$ represents an input image and $\bY_i = \{\by_{i}^{1}, ..., \by_{i}^{K}\}$ a corresponding multi-task label with ground-truth data for each one of $K$ modalities. The most standard way to perform AL is to iteratively go through the following four steps shown in Fig.~\ref{fig:vanilla-mtal}:
\begin{enumerate}
  \item Train the neural network using all available samples and all of the modalities in $\bL$.
  \item Estimate the prediction uncertainty for each image in $\bU$.
  \item Provide ground-truth data for the images in $\bU$ whose prediction uncertainty is highest.
  \item Remove the newly labelled samples from $\bU$ and add them to $\bL$.
  \label{en: acl}
\end{enumerate}
These labelling and training steps can be repeated until a stopping condition is met, such as reaching a predetermined model accuracy or labelling budget.

\begin{figure}[h!]
    \centering
    \includegraphics[width=0.47\textwidth]{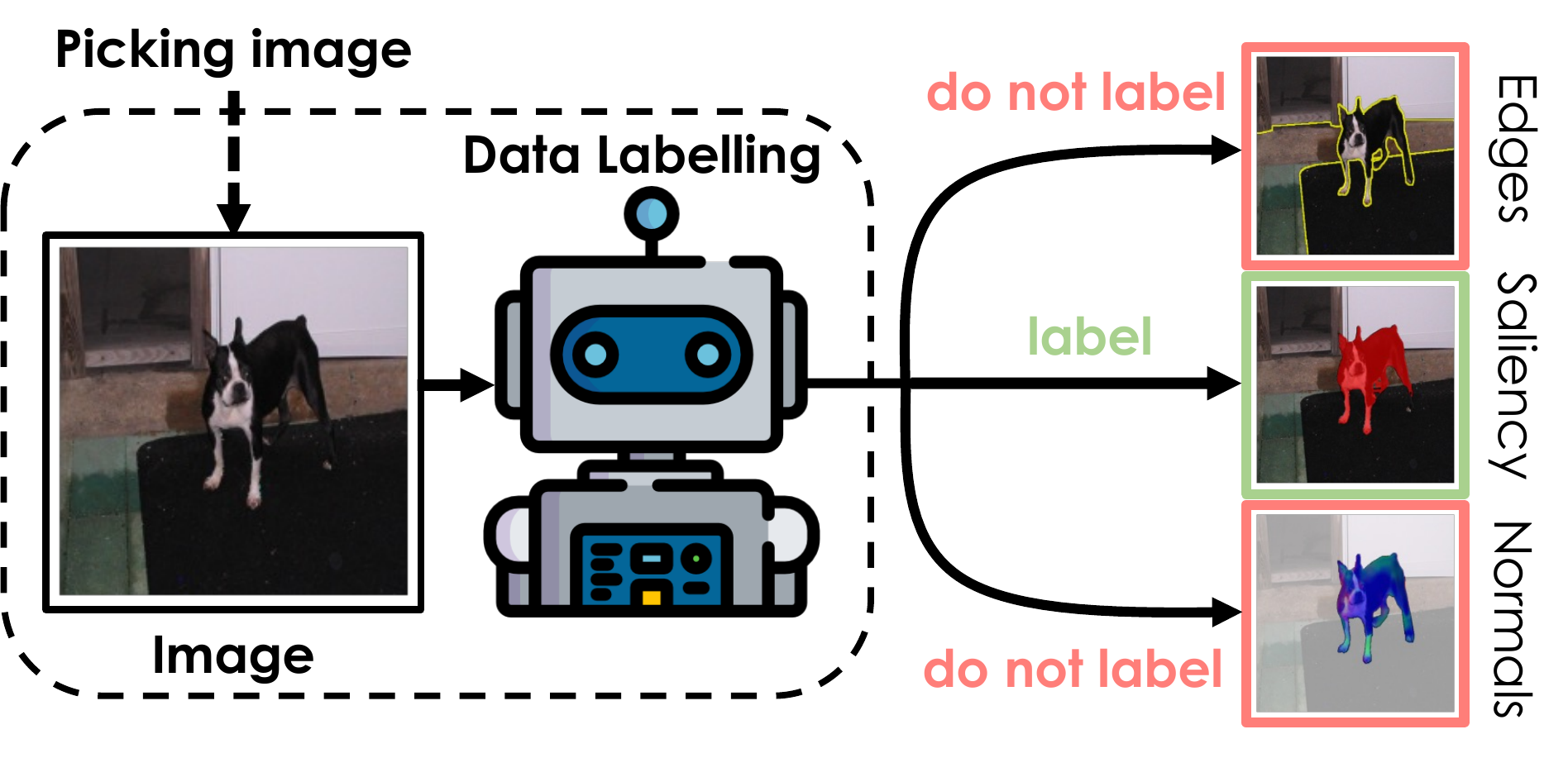}
    \vspace{-5mm}
    \caption{\small {\bf Partial labelling.} In contrast to conventional active learning methods, \pale{} chooses both which image and which modalities to label. Hence, it selects image / modality pairs from  the unlabelled pool.}
    \label{fig:labelling}
\end{figure}

At the heart of \pale{} are two key changes to this vanilla approach. 
\begin{enumerate}

  \item At each training step, we provide only a subset of $\{\by_{i}^{1}, ..., \by_{i}^{K}\}$  for each $\bx_i$ we chose to annotate, as shown in Figs.~\ref{fig:mtask} and \ref{fig:labelling}.  During subsequent steps, the tasks for which no ground-truth data is provided are simply ignored. The benefit is that we then need to provide far fewer annotations and that each one has a greater impact. As a result, when using the same total number of annotations, our networks end up being better trained. 
  
  \item When estimating the uncertainty, we rely on the distillation stage of PAD-Net to use the available ground-truth labels $\by_{i}^{k}$  to help predict the others. This helps find the most informative samples and further boosts performance.
  
\end{enumerate}
We use a modified PAD-Net architecture~\cite{Xu18c} to implement \pale{} because it comprises a  Multi-Modal Distillation block (MMD) that enforces consistency between modalities during training, as shown in Fig.~\ref{fig:padnetwork}. We first use a concrete example to motivate our approach and then describe its two key components.


\subsection{Motivating Example}

Let us consider an image dataset along with associated ground-truth depth and normals maps. Since the ones can be computed analytically from the others, depths and normals are correlated. Let us consider a setup where $\bL$ consists of $N_{L}$ labelled samples (image, depth, normals) and $\bU$ of $N_{U}$ unlabelled samples (only images). In a standard AL setup,   let us assume that we can only afford to label half of the data in $\bU$. Hence, we would end up supplying 2 annotations---normal and depth---per image in $\bU$ for a total of $N_{U}$ labels. 

By contrast,  \pale{} can use the MMD of PadNet to predict one label from the other. Thus, given  $N_{U}$ labels, we can infer automatically an additional  $N_{U}$ and use $2 N_{U}$ labels to learn the weights instead of only  $N_{U}$, at no extra annotation cost. In practice, some of the inferred labels will be less accurate than the ground-truth ones. Nevertheless, our experiments show that the benefit remains. We now turn to the actual implementation of our partial annotation scheme.

\subsection{Partial Annotations}
\label{sec:choosing}

\begin{figure}[h]
\centering
\begin{tabular}{@{}c@{}c@{}}
  \includegraphics[width=0.22\textwidth]{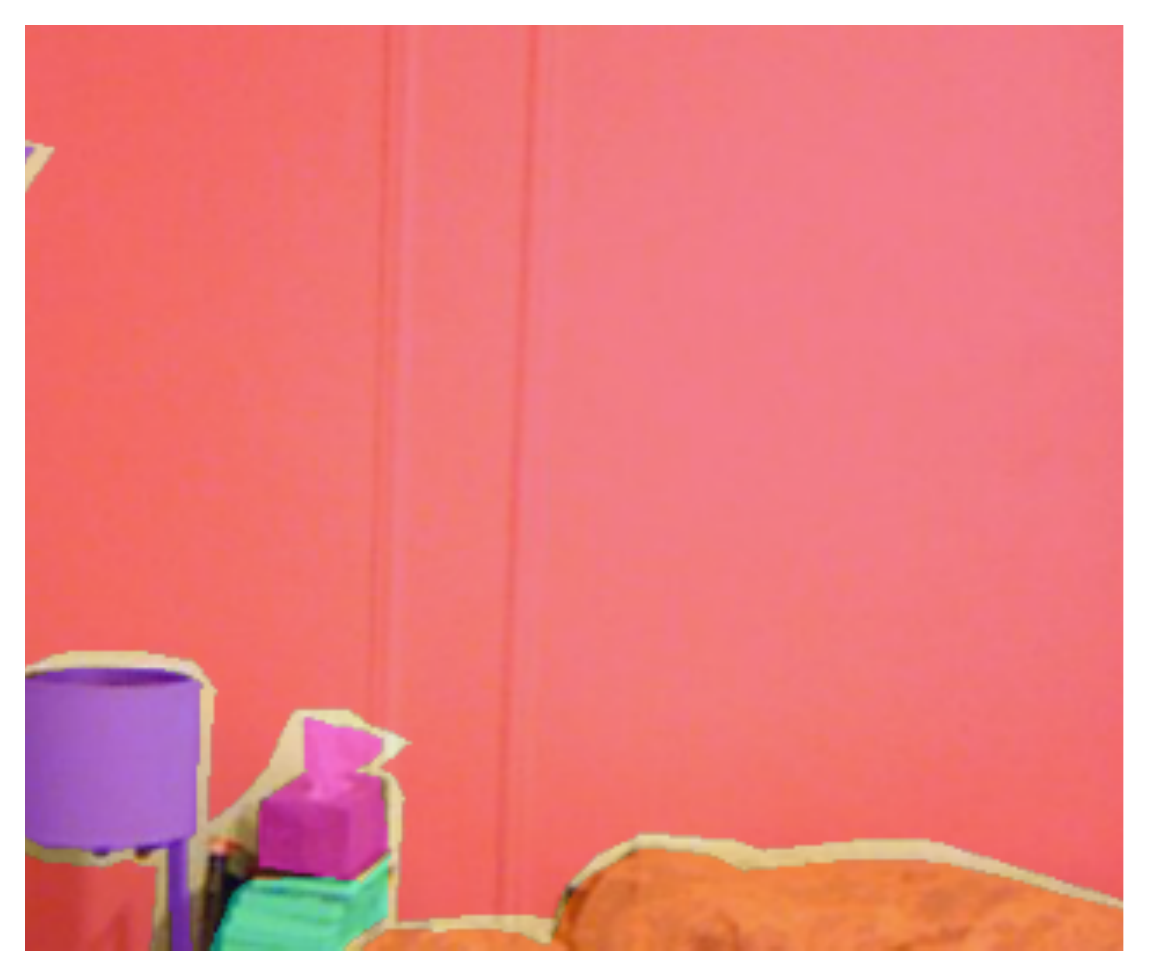} &   
  \hspace{3mm}
  \includegraphics[width=0.22\textwidth]{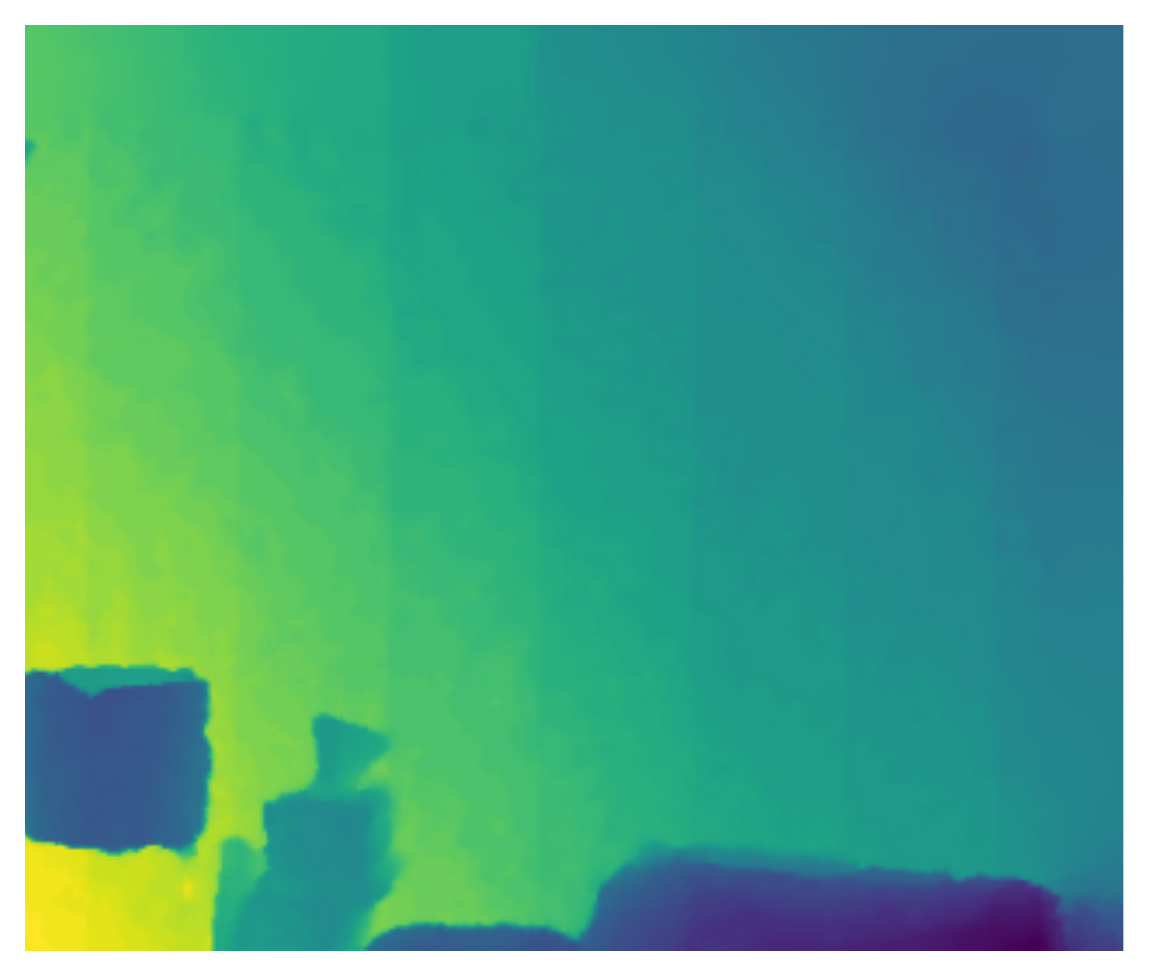} \\
\end{tabular}
\vspace{-3mm}
\caption{\small
\textbf{Picking the most difficult targets.} This plot depicts two targets for one image: semantic segmentation (\textbf{left}) and depth map (\textbf{right}). The segmentation task is relatively simple since a major part of the image is occupied by the wall. At the same time, mostly textureless and monochrome wall makes it harder to make depth prediction. This way, the depth estimation task turns out to be more complicated for our model for this particular image. 
}
\label{fig:diff-targets}
\end{figure}

As shown in Figures~\ref{fig:labelling} and~\ref{fig:diff-targets},  we aim to select the most uncertain image~/~modality pairs $(\bx_i,\by_i^k)$ for annotating, as opposed to the most uncertain image for which we would provide labels for all modalities, as in traditional methods. This amounts to making such methods more flexible. The algorithm can still request labels for all modalities if it is likely to be beneficial but it does not have to. 


As in many other approaches, we rely on uncertainty sampling~\cite{Lewis94} to select image / modality pairs. In practice, we use MC-Dropout~\cite{Gal16a} to estimate individual uncertainties for \textit{each} of the targets and the resulting estimates are not always consistent across modalities, which makes comparisons unreliable. We address this problem by normalizing uncertainties using the following min-max scheme. 

Let us consider a network trained on partially labelled data. For the $N$ current images from $\bU$ and all modalities, it produces uncertainty estimates $u_i^k$ for $1 \leq i \leq N$ and $1 \leq k \leq K$. We normalize the uncertainties into the interval $[0,1]$ by writing 
\begin{align}
\widetilde{u}_i^k &= \frac{u_i^k-u_{\rm{min}}^k}{ u_{\rm{max}}^k-u_{\rm{min}}^k} \ ;  , \\ 
\mbox{ where }u_{\rm{min}}^k = &\min_i u_i^k \mbox{ and } u_{\rm{max}}^k = \max_i u_i^k \; .\nonumber
\end{align}
In practice, $u_{\rm{min}}^k$ and $u_{\rm{max}}^k$ are computed at the first active learning step and reused for all others. This makes normalized uncertainties comparable across iterations. 

A difficulty we face when implementing this scheme is that some of the tasks we deal with are categorical while others are continuous. In practice, categorical uncertainties can be estimated using Shannon Entropy~\cite{Shannon48} expressed as 
$$H(P):= -\sum_{i} p_{i} \log{p_{i}} \; ,$$ 
and uncertainties on continuous variables as differential entropy
$$h(p) := -\int_x p(x)\log\left(p(x)\right) dx \; . $$
These two entropies are not directly comparable with each other. However, we show in the supplementary material that differential entropies can be approximated by categorical ones. Thus the min-max scheme introduced above to make them comparable makes theoretical sense. 

\subsection{Partial Labels at Inference Time}
\label{sec:partial_inference}

As discussed in Section~\ref{sec:related}, PAD-Net~\cite{Xu18c} is an effective architecture for multiple visual tasks. As shown in Fig.~\ref{fig:padnetwork}, it yields initial predictions for each task and then merges them into final ones and enforces consistency. In \pale{}, we replace the initial predictions for some modalities by ground-truth values when they happen to be provided. In other words, we reuse known values for some tasks to help accomplish the others. 

In the result section we demonstrate empirically that this provides a significant performance boost. Here we offer a theoretical justification. To this end, we rely on Fisher information~\cite{LeCam12}. Let  $\mathcal{I}_{\by}$ be the amount of information gained by sampling random variable $\by$ once. 
Let us consider two correlated random variables $\by_1$ and $\by_2$. We want to extract as much information about them by picking $2N$ samples. One way is to sample each one $N$ times in parallel for a Fisher information gain of $N\mathcal{I}_{\by_1,\by_2}$, which is essentially what a normal AL procedure does. Alternatively, we could also sample $\by_1$  $N$ times given the ground-truth value of $\by_2$ and gain $N\mathcal{I}_{\by_1 | \by_2}$. Assuming we can model conditional probabilities well enough, 
it does not matter what value of $\by_2$ we use. We can then reverse the roles for a total information gain of $N\mathcal{I}_{\by_1 | \by_2} + N\mathcal{I}_{\by_2 | \by_1}$, which is essentially what we do when we use one label to predict another. Using chain rule decomposition for Fisher information, we write
\begin{align}
N\mathcal{I}_{\by_1,\by_2} = N\mathcal{I}_{\by_1} + N\mathcal{I}_{\by_2 | \by_1} 
                                            \leq  N\mathcal{I}_{\by_1 | \by_2} +  N\mathcal{I}_{\by_2 | \by_1} \label{eq:fisher}
\end{align}
because $N\mathcal{I}_{\by_1} \;\; \le \;\; N\mathcal{I}_{\by_1 | \by_2}$. This holds because, if $\by_1$ and $\by_2$ are correlated and we know the value of $\by_2$, it becomes easier to say what the value of $\by_1$ should be.

Applying this line of reasoning to a standard active learning setup, when we pick $N$ samples and label all modalities, we provide the information $N\mathcal{I}_{\by_1,\by_2 | \bx} $. By contrast, in our partial labelling scheme, we provide $N\mathcal{I}_{\by_1 | \by_2, \bx} +  N\mathcal{I}_{\by_2 | \by_1, \bx} $, which in general is more according to Eq.~\ref{eq:fisher}.


\section{Experiments}

In this section, we validate our approach and we provide additional details in the supplementary material. 

\subsection{Baselines}

As discussed in Section~\ref{sec:related}, multi-task active learning in a deep learning context has received little attention. Hence, for comparison purposes, we define the four following baselines inspired by the few existing techniques for multi-task active learning that exist but are not deep and popular single-task AL methods.

\begin{itemize}

 \item \textit{Random.} The simplest possible approach where new samples are picked at random during the active learning stage, without reference to uncertainty or trying to determine what samples should be labelled first.

 \item \textit{Ranking-Based Active Learning (RBAL).} As in~\cite{Reichart08}, we compute an uncertainty value for each sample / modality pair and rank those uncertainties within their particular modality. Then, for each image $i$ and modality $k$, the pair $(\bx_i,\by_i^k)$ is assigned rank $r_{i}^{k}$. We use the combined rank $R_{i} = \sum_{k=1}^{K} r_{i}^{k}$ to select the samples to be considered first. For these samples, we provide labels for all modalities, again as in~\cite{Reichart08}.
 
 \item \textit{Learning Loss for Active Learning (LLoss)}: As in~\cite{Yoo19}, we use an additional model head to estimate not only a desired value but also a likely deviation from that value. At training time, ground-truth deviations are available and the network can be trained in the usual way.  For sample selection purposes, we estimate the deviation for each image $\bx_i$ in the unlabelled pool and pick the hardest examples according to it. 
  
 \item \textit{Core-Set Active Learning (CS)}: As in~\cite{Sener17}, given a trained model $\mathcal{M}$ we extract deep features $\bz_{i}$ for each image $\bx_{i}$ in the unlabelled pool. Then we run the core-set solver from~\cite{Sener17}  to find a set of samples to label such that when we learn a model, the performance of the model on the labelled subset and that on the whole dataset are as close as possible.

\end{itemize}
We compare these baselines to our own \pale{} approach on the visual multi-task datasets described below and use the same training and evaluation setups as in~\cite{Vandenhende21}.

\subsection{Metrics}
\label{sec:metrics}

Since some tasks involve categorical targets and others continuous ones, we use different metrics to gauge how good our trained networks are at these various tasks. As in \cite{Vandenhende21}, we use the following. 
\begin{itemize}

 \item \textit{Mean Intersection Over Union (mIoU) } ($\uparrow$): IoU is a common evaluation metric for semantic segmentation and detection tasks. We compute the IoU for each individual class and report the mean of these values over all classes. 
 
 \item \textit{Optimal Dataset F-measure (odsF)} ($\uparrow$): Originally proposed in~\cite{Martin04}, it casts pixel-wise detection as a classification task and computes precision/recall curves. Given these, we calculate the best F-measure (or F-score) on the estimated PR-curve. 
 
 \item \textit{Mean Angle Error (mErr)} ($\downarrow$): It is used to evaluate the distance between vectors. The dot products between vectors that have been normalized are computed and averaged over all vector pairs. 
 
 \item \textit{Root-Mean-Square Error (RMSE)} ($\downarrow$): It is widely used for regression tasks and is defined as the square root of the mean square error. 
 
\end{itemize}

\subsection{Results on the NYUv2 Dataset}
\label{sec:nyud}

We use the same experimental protocol as in~\cite{Vandenhende21}. Hence, we use a PAD-Net~\cite{Xu18c} with  an HRNet~\cite{Wang20i} backbone.  Training details and hyper-parameters are given in the supplementary material. 

\begin{figure}[h]
\centering
\begin{tabular}{@{}c@{}c@{}}
  \includegraphics[width=0.24\textwidth]{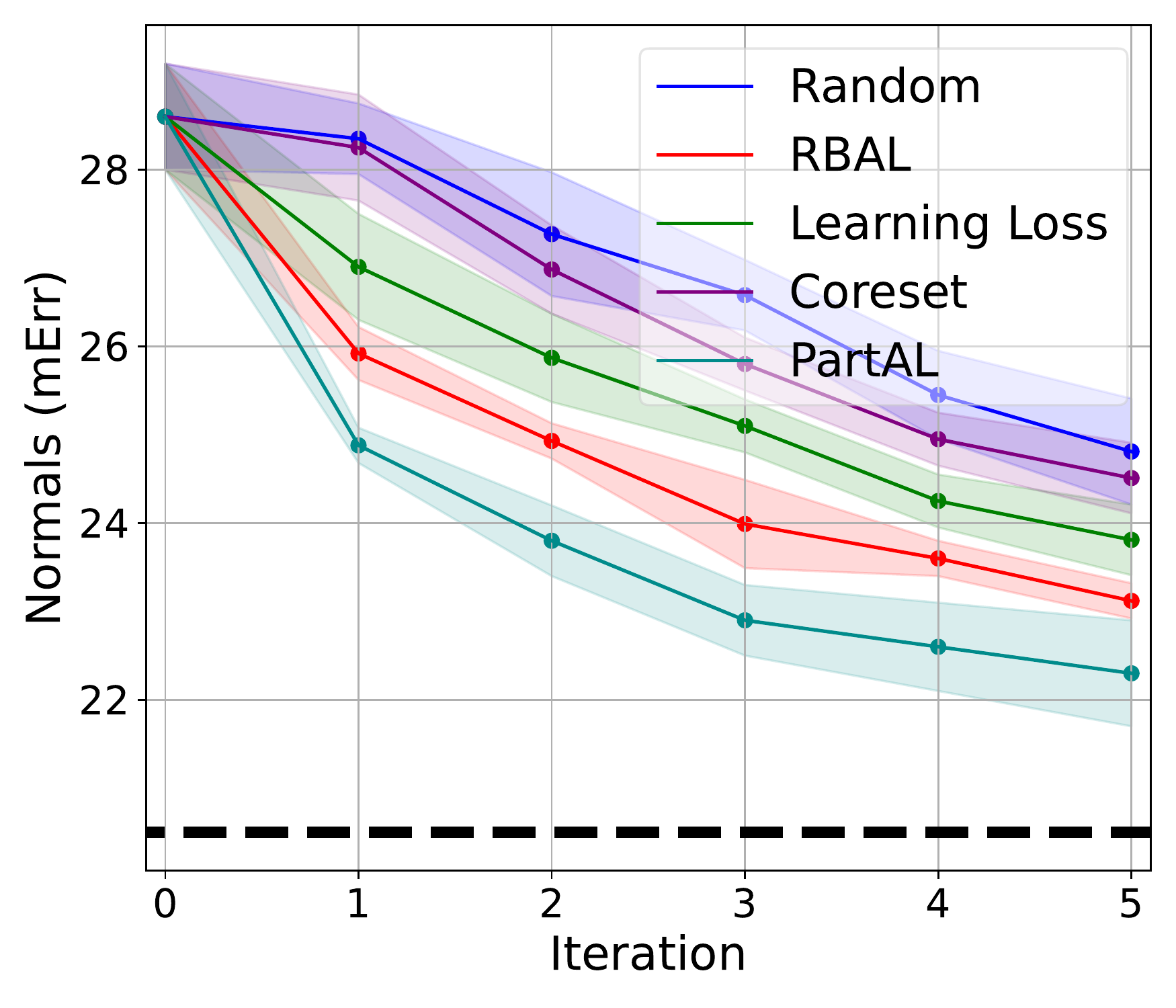} &    \includegraphics[width=0.24\textwidth]{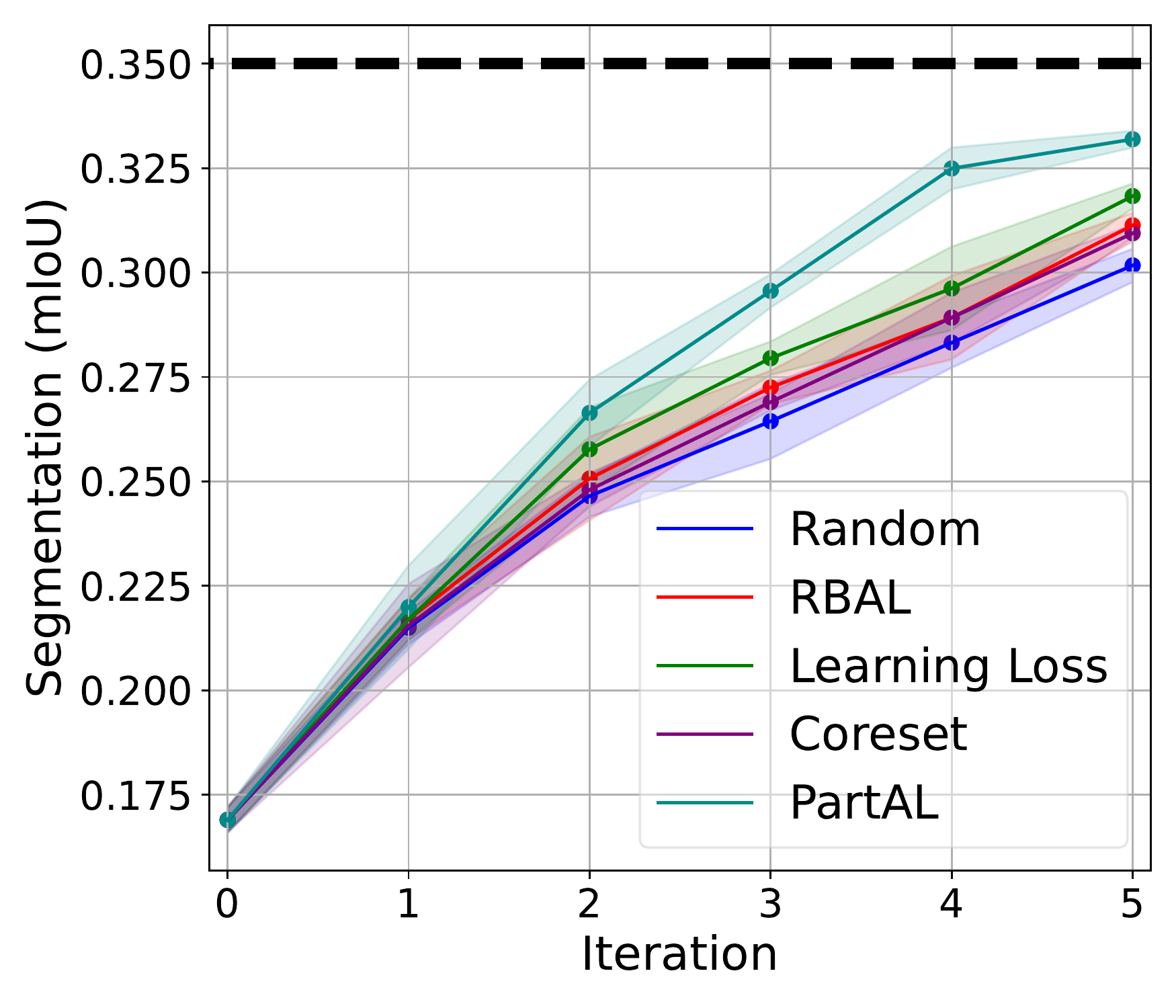} \\[-2mm]
   \hspace{6ex}(a) & \hspace{6ex}(b) \\
   \includegraphics[width=0.24\textwidth]{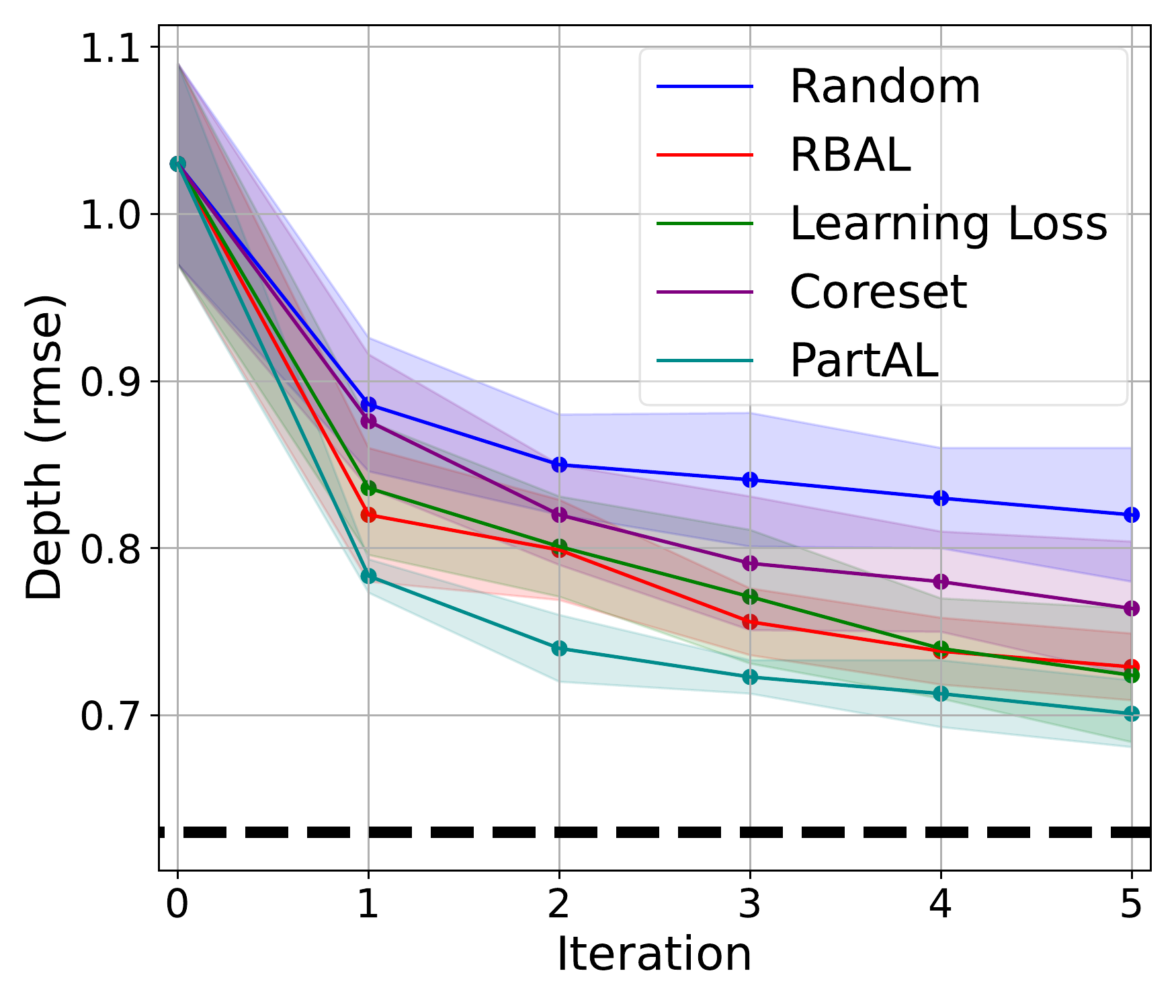} &  \includegraphics[width=0.24\textwidth]{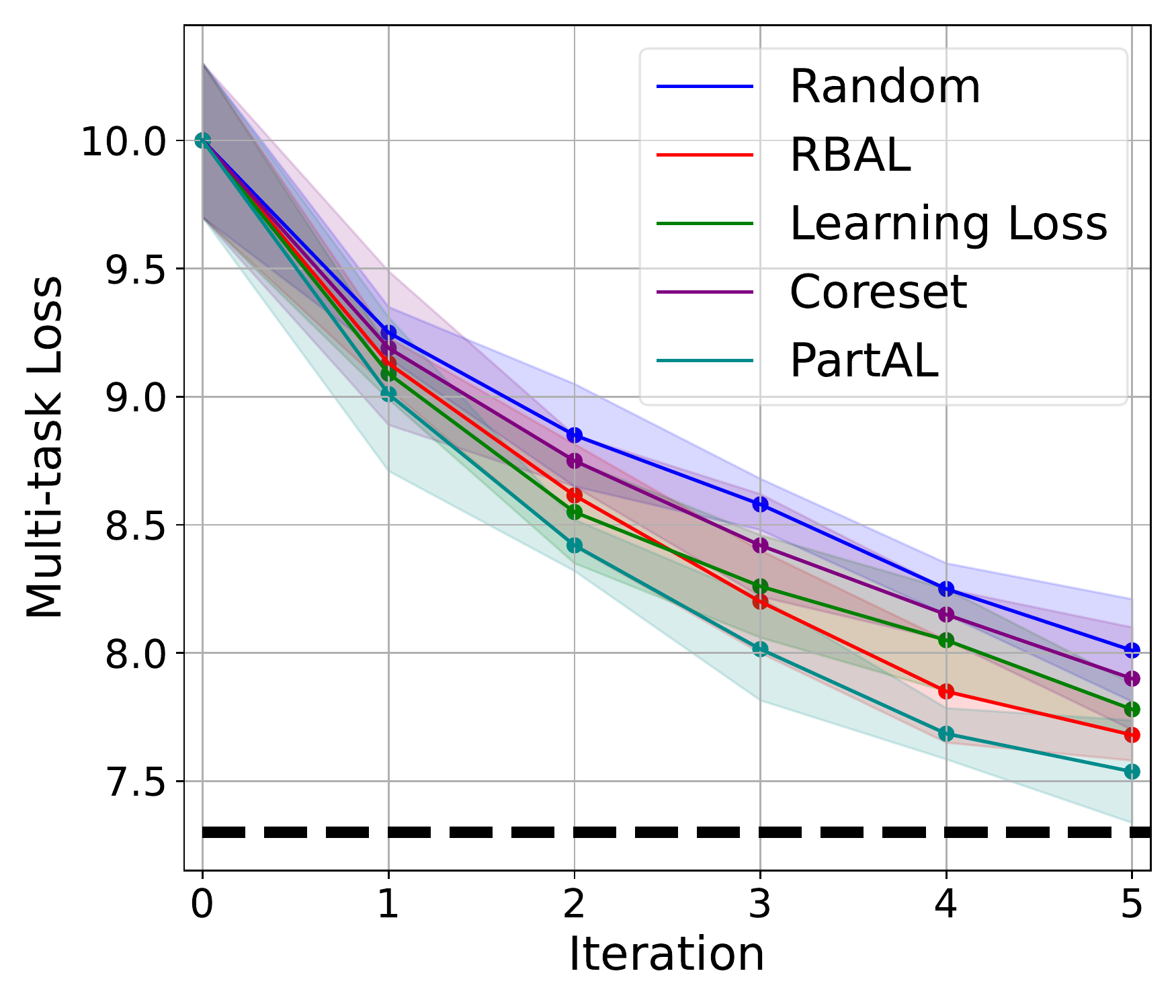} \\[-2mm]
   \hspace{6ex}(c) & \hspace{6ex}(d) \\
\end{tabular}
\vspace{-3mm}
\caption{\small
\textbf{Evaluation on NYUDv2.} We ran each active learning procedure 3 times with different random seeds. Each line represents a metric computed on the evaluation set as a function of the number of AL iterations. They are color-coded according to the method used to perform the procedure: \textcolor[HTML]{0a0afb}{\textbf{\textit{Random}}}, \textcolor[HTML]{f51214}{\textbf{\textit{RBAL}}}, \textcolor[HTML]{830c83}{\textbf{\textit{Coreset}}}. \textcolor[HTML]{198d19}{\textbf{\textit{Learning Loss}}}, and
\textcolor[HTML]{088e8e}{\textbf{\textit{PartAL}}}. The  shaded  areas denote the corresponding standard deviations. The \textcolor[HTML]{088e8e}{\textbf{\textit{PartAL}}} curve consistently decreases faster than the others.
\textbf{(a)}  mErr for normals. \textbf{(b)} mIoU for semantic segmentation. \textbf{(c)} RMSE for depth estimation. \textbf{(d)} Multi-task error.}
\label{fig:nyu}
\end{figure}

NYUDv2 consists of indoor images with annotated depth and segmentation maps~\cite{Silberman12}. As in~\cite{Vandenhende21,Xu18c}, we use normals estimated from depths as additional labels. For all methods, we start with $100$ images with all labels provided, that is depth, segmentation, and normals. At each further active learning step, we add $80$ new images with ground-truth labels for each of the three modalities for \textit{Random}, \textit{RBAL}, \textit{Coreset} and \textit{Learning Loss}, and $240$ labels chosen among all ground-truth labels of all images for \pale{}. In short, we supply the same number of labels to all methods. In Fig.~\ref{fig:nyu}(a,b,c), we report the errors made on the test set as a function of the number of AL iterations for normals, segmentation, and depth as a function of the number of AL iterations. In Fig.~\ref{fig:nyu}(d) we similarly plot curves that represent the cumulative multi-task error computed as an aggregated metric for all targets, as described in more detail in the supplementary material. The \pale{} error curves are consistently below the others. 

We also provide results with full supervision using {\it all} the training data.  They are depicted by thick dashed lines in all four plots of Fig.~\ref{fig:nyu} and closely match those reported in~\cite{Vandenhende21}. To quantify this, after the last AL iteration, we computed the difference $\delta$ between the metrics we obtain using the various AL methods discussed above and those we obtain with full supervision. At this point, we only use about $60\%$ of the available training data and our performance is close to what full supervision delivers.


\begin{table}[ht]
    \begin{center}
      \begin{tabular}{c|ccccc} 
         & Random & RBAL & CS & LLoss & PartAL  \\
        \hline
        Depth $\delta$ & 0.19 & 0.1 & 0.13 & 0.09 & \textbf{0.07} \\
        Segm. $\delta$ & 4.8 & 3.8 & 4.0 & 3.5 & \textbf{1.8}\\
        Norm. $\delta$ & 4.30 & 2.62 & 4.01 & 3.31 & \textbf{1.8}\\
      \end{tabular}
    \end{center}
    \caption{\small {\bf NYUDv2 final performance.} Final depth estimation, normals estimation and semantic segmentation deltas for a model trained on final data acquired by corresponding method. After 5 iterations of active learning, our method produced the best performance.} 
      \label{tab:nyu}  
      \vspace{-0.3cm}
  \end{table}

\subsection{Results on the PASCAL Dataset}
\label{sec:pascal}

The PASCAL dataset~\cite{Everingham10} is a popular benchmark for dense prediction tasks. As in~\cite{Vandenhende21}, we use the PASCAL-Context split~\cite{Chen14a} that has annotations for semantic segmentation, human part segmentation and edge detection. In addition, we consider surface normal estimation and saliency detection. For all methods, the initial set consists of fixed 1000 labelled images with five  modalities each (normals, segmentation, depth, human segmentation, and saliency). At each AL iteration, \textit{Random}, \textit{RBAL}, \textit{Coreset} and \textit{Learning Loss} query 300 fully-labelled images and \pale{} queries 1500 image / target pairs. Again, in the end, all methods use the same labelling budget. 

\begin{figure}[h]
\centering
\begin{tabular}{@{}c@{}c@{}}
  \includegraphics[width=0.24\textwidth]{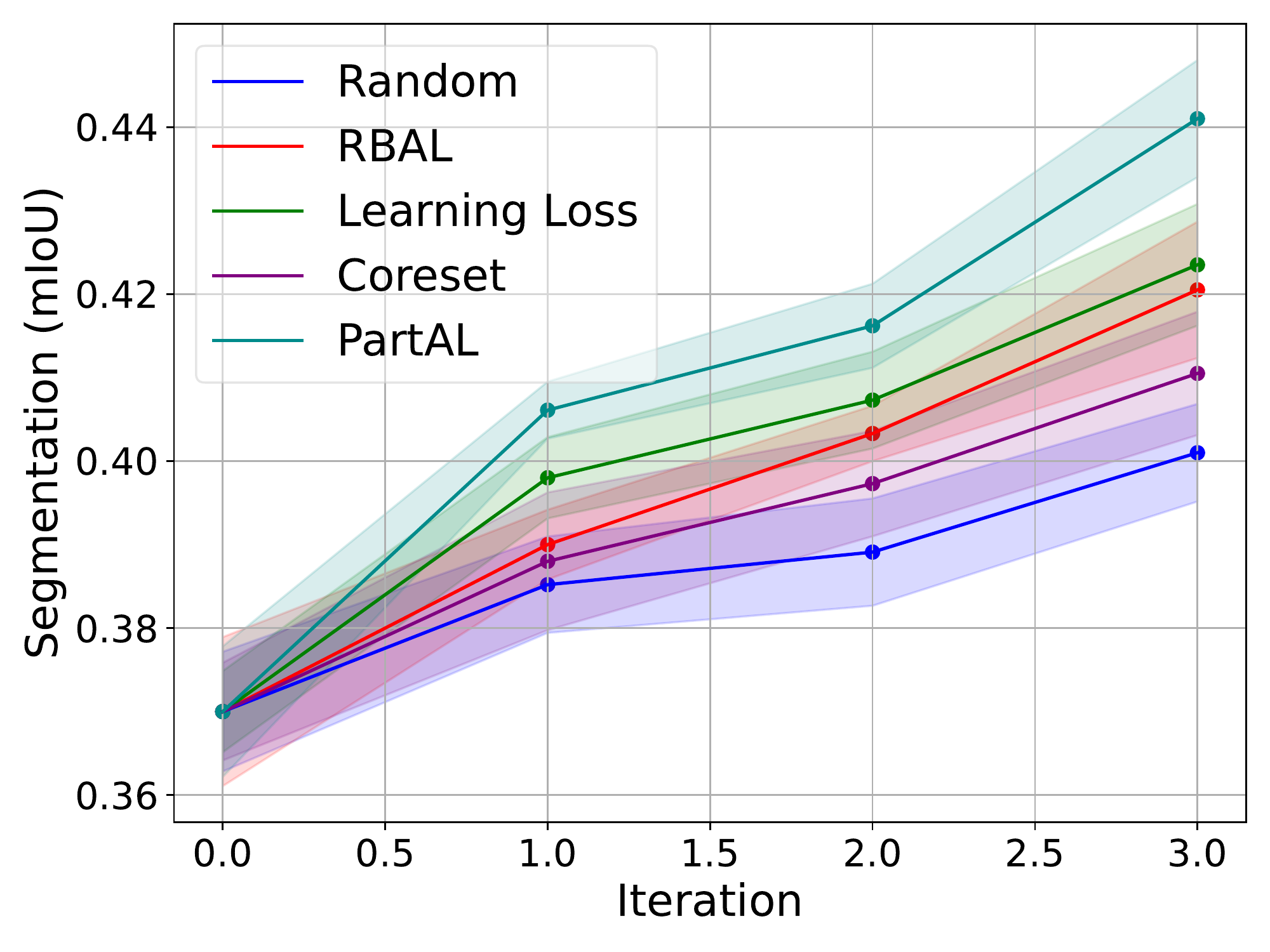} &    \includegraphics[width=0.24\textwidth]{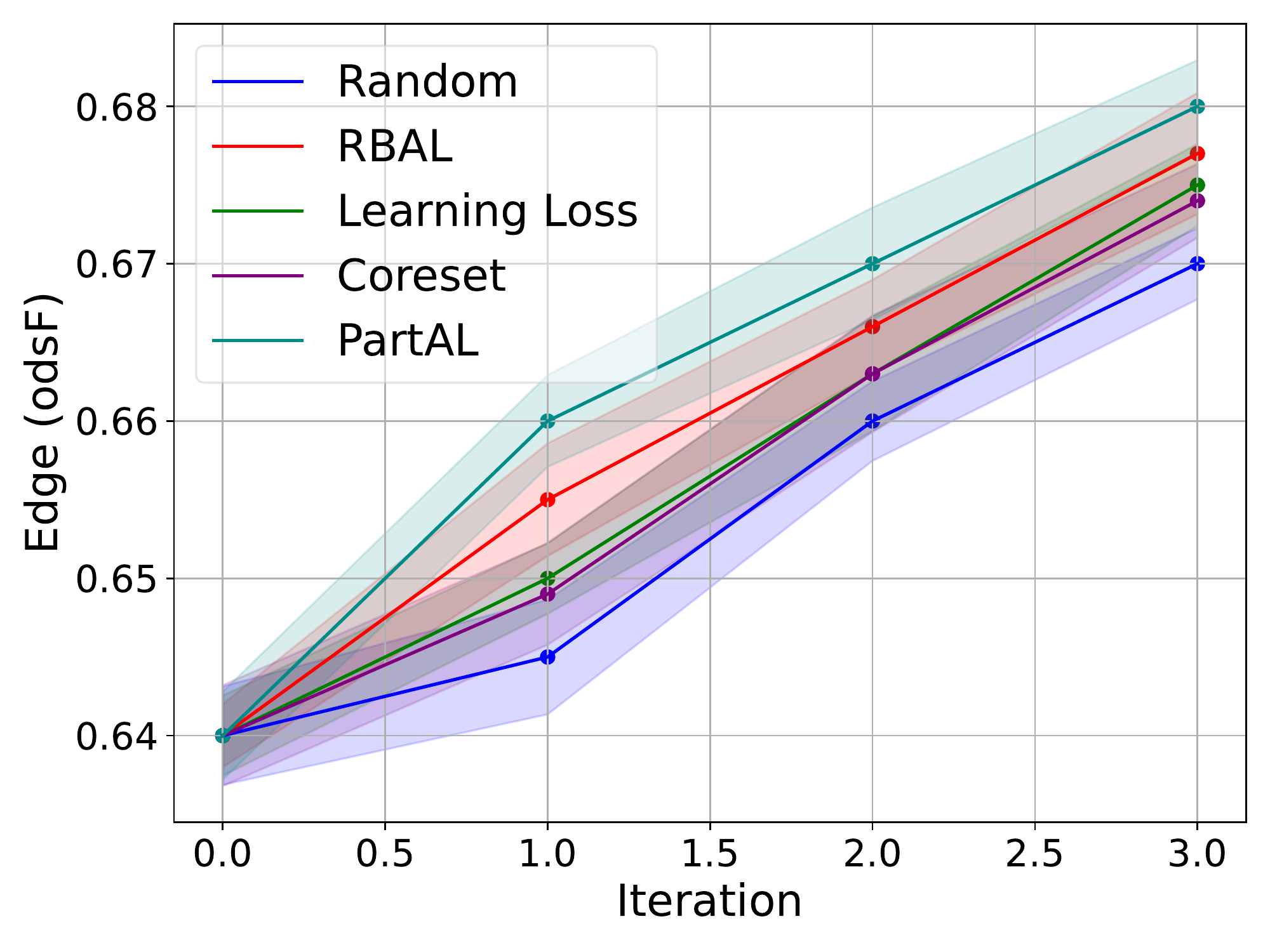} \\[-2mm]
   \hspace{6ex}(a) & \hspace{6ex}(b) \\
   \includegraphics[width=0.24\textwidth]{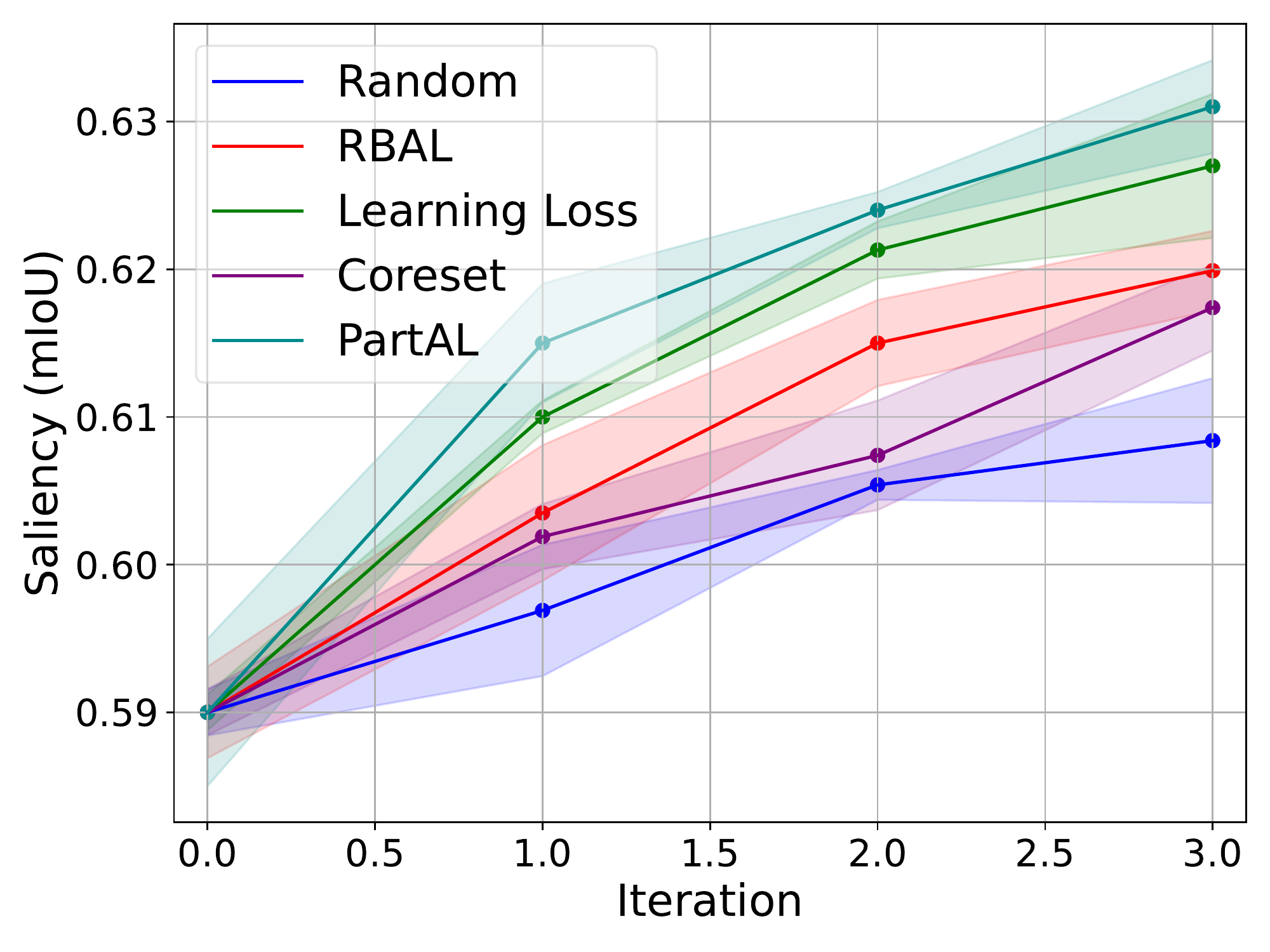} &  \includegraphics[width=0.24\textwidth]{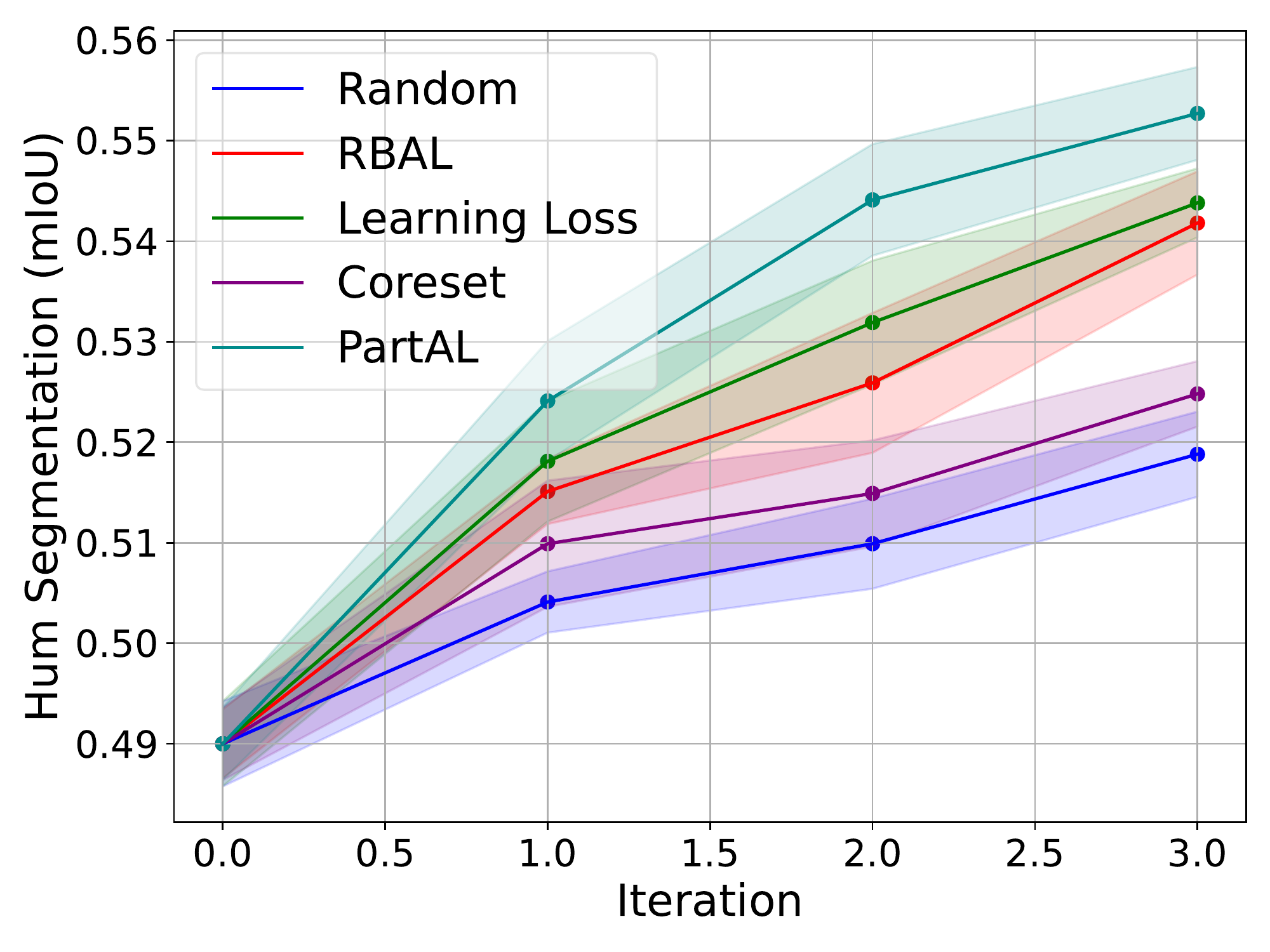} \\[-2mm]
   \hspace{6ex}(c) & \hspace{6ex}(d) \\
   \includegraphics[width=0.24\textwidth]{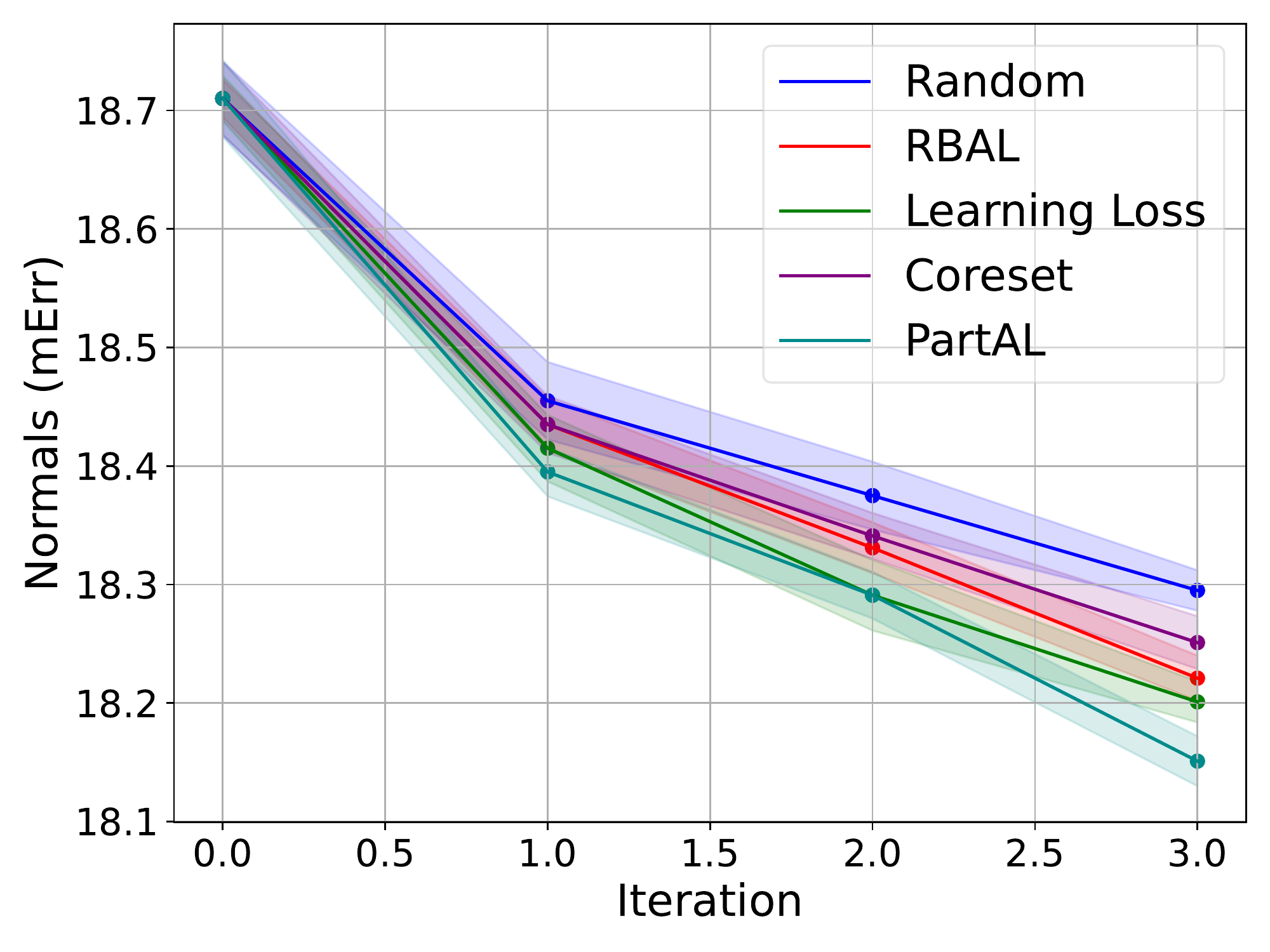} &  \includegraphics[width=0.24\textwidth]{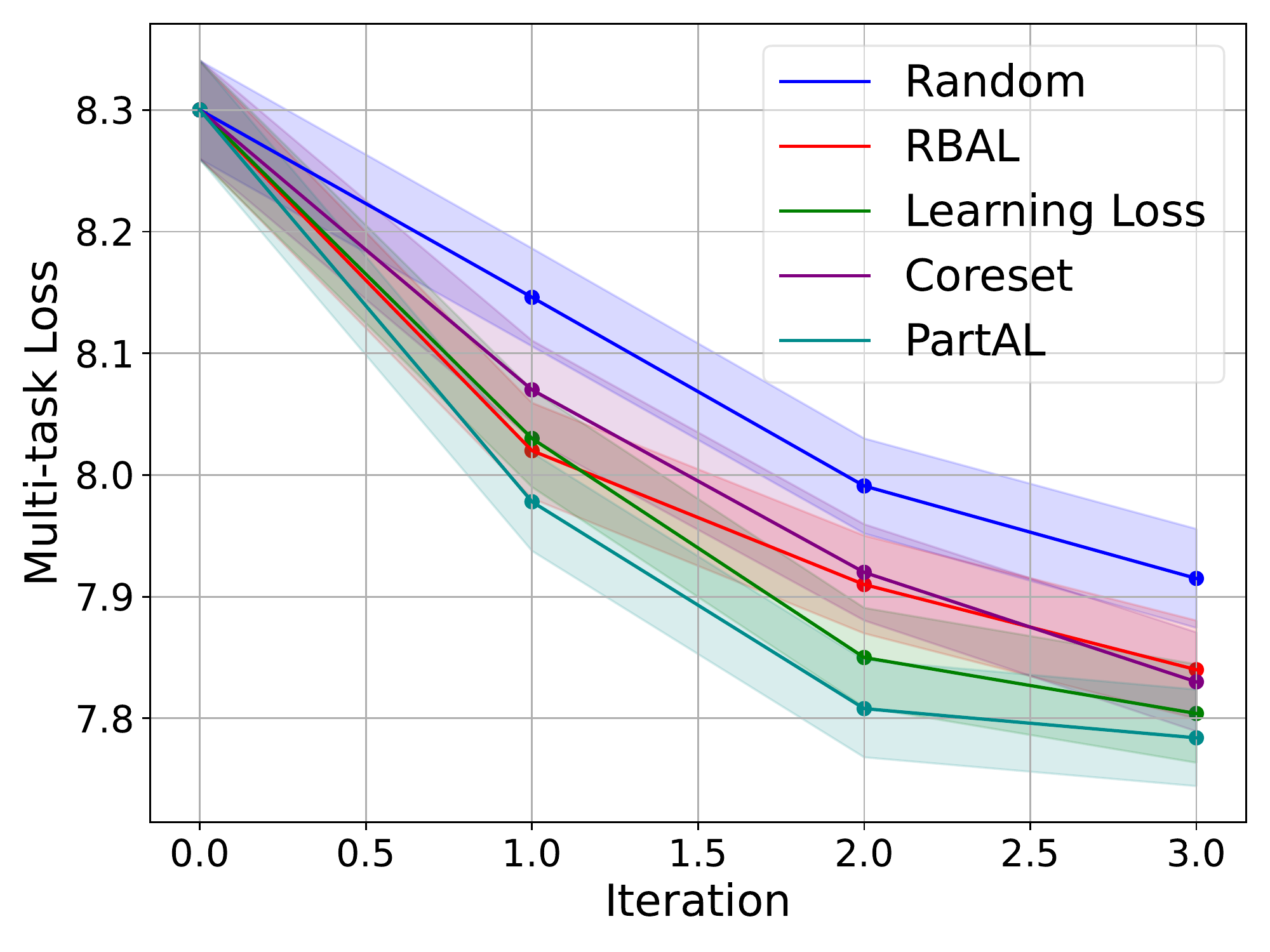} \\[-2mm]
   \hspace{6ex}(e) & \hspace{6ex}(f) \\   
\end{tabular}
\vspace{-3mm}
\caption{\small
\textbf{Evaluation on PASCAL.}
We ran each active learning procedure 3 times with different random seeds. As in Fig.~\ref{fig:nyu}, each color-coded curve---\textcolor[HTML]{0a0afb}{\textbf{\textit{Random}}}, \textcolor[HTML]{f51214}{\textbf{\textit{RBAL}}}, 
\textcolor[HTML]{830c83}{\textbf{\textit{Coreset}}}, \textcolor[HTML]{198d19}{\textbf{\textit{Learning Loss}}}, and \textcolor[HTML]{088e8e}{\textbf{\textit{PartAL}}}---represents test error as a function of the number of AL iterations.  \textbf{(a)} mIoU for semantic segmentation, \textbf{(b)} odsF for edge detection \textbf{(c)} mIoU for saliency estimation, \textbf{(d)} mIoU for human parts segmentation, \textbf{(e)} mErr for normals estimation, \textbf{(f)} total multi-task loss. Once again, \textcolor[HTML]{088e8e}{\textbf{\textit{PartAL}}} exhibits the best performance. 
}
\label{fig:pascal}
\end{figure}

As for the NYUv2 experiments, Fig.~\ref{fig:pascal} depicts the behavior of each method as a function of the number of AL iterations. In Tab.~\ref{tab:pascal}, we report $\delta$ values similar to those in the NYU experiment, with all the AL approaches using about $40\%$ of available training data. \pale{} again consistently outperforms the other approaches. 


\begin{table}[ht]
    \begin{center}
      \begin{tabular}{c|ccccc} 
         & Random & RBAL & CS & LLoss & PartAL  \\
        \hline
        Sal. $\delta$ & 5.0 & 3.8 & 4.1 & 3.1 & \textbf{2.7} \\
        HSegm. $\delta$ & 7.7 & 5.4 & 7.1 & 5.2 & \textbf{4.3} \\
        Segm. $\delta$ & 13.5 & 11.6 & 12.6 & 11.3 & \textbf{9.5}\\
        Norm. $\delta$ & 3.00 & 2.92 & 2.95 & 2.90 & \textbf{2.85}\\
        Edge $\delta$ & 5.50 & 4.80 & 5.10 & 5.00 & \textbf{4.50}\\
      \end{tabular}
    \end{center}
    \vspace{-0.5cm}
    \caption{\small {\bf PASCAL final performance.} Final depth estimation, normals estimation and semantic segmentation results for a models trained on final data acquired by corresponding method. After 5 iterations of active learning, our methdos produced the best performance.} 
      \label{tab:pascal}  
      \vspace{-0.3cm}
  \end{table}


\subsection{Ablation Study}

As discussed above, our contribution is two-fold. We propose a partial labelling scheme at training time and a way to reuse available ground-truth labels at inference time. We now study their individual influence. 


\begin{figure}[h!]
    \centering
    \begin{tabular}{@{}c@{}c@{}}
      \includegraphics[width=0.24\textwidth]{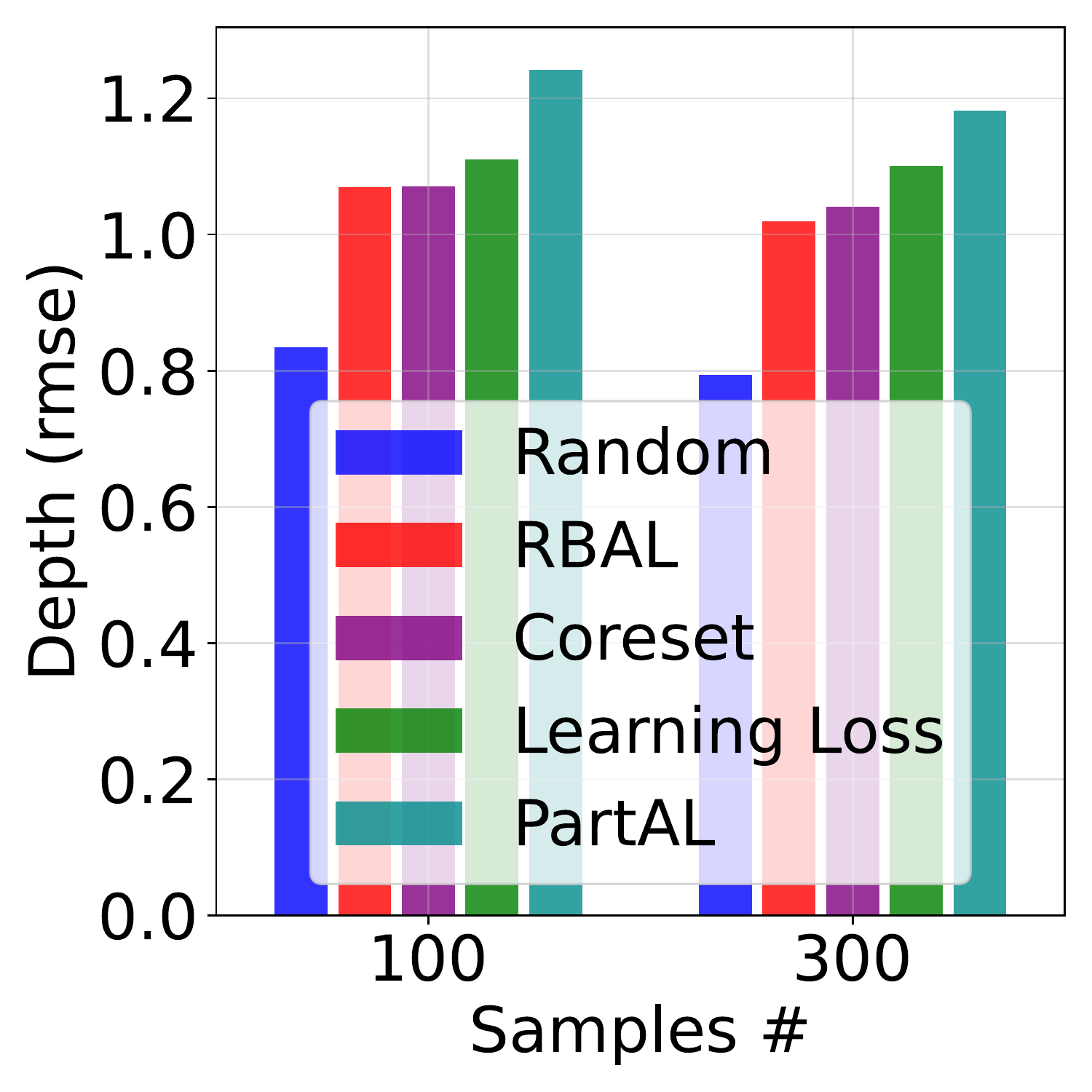} & 
      \includegraphics[width=0.24\textwidth]{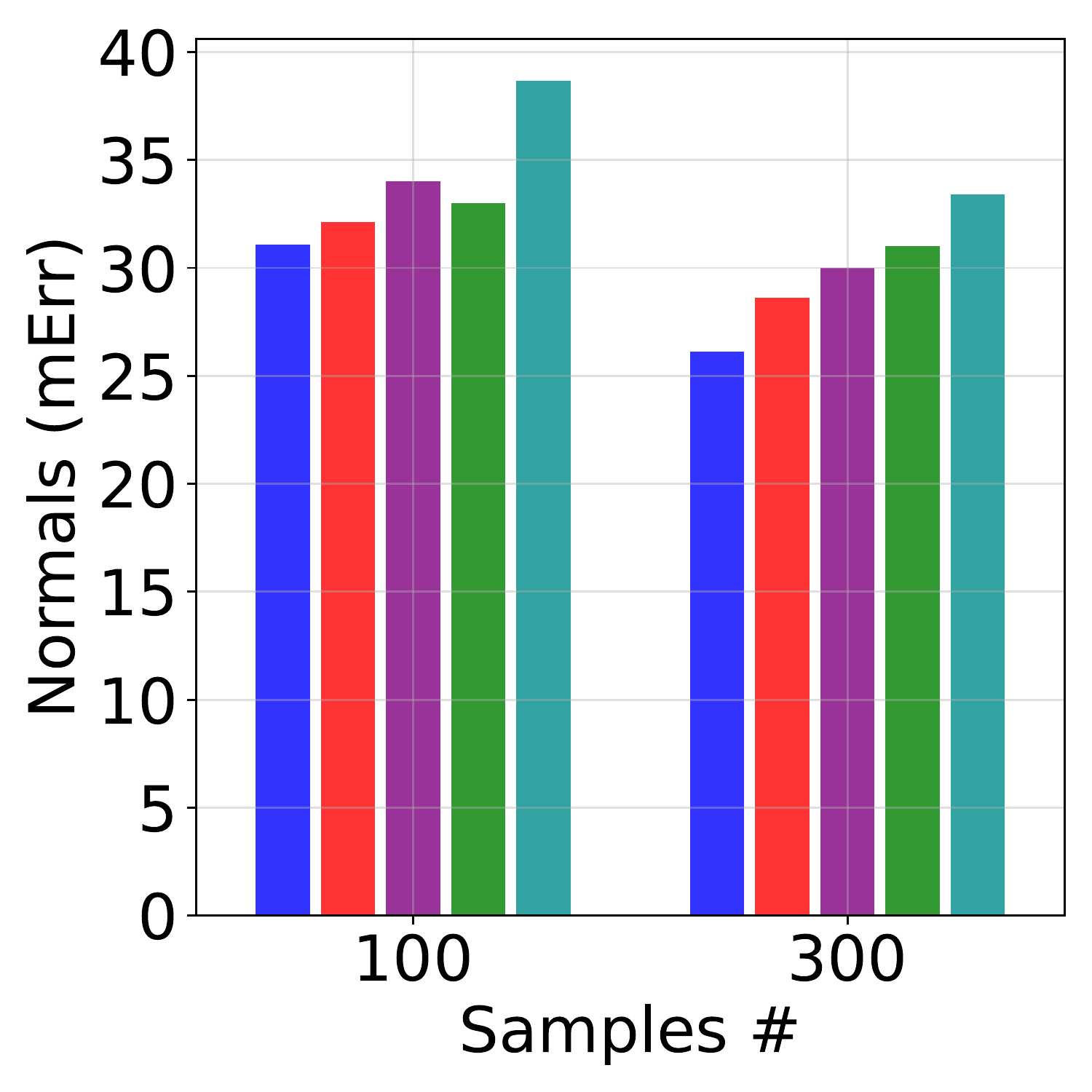}\\
      \includegraphics[width=0.24\textwidth]{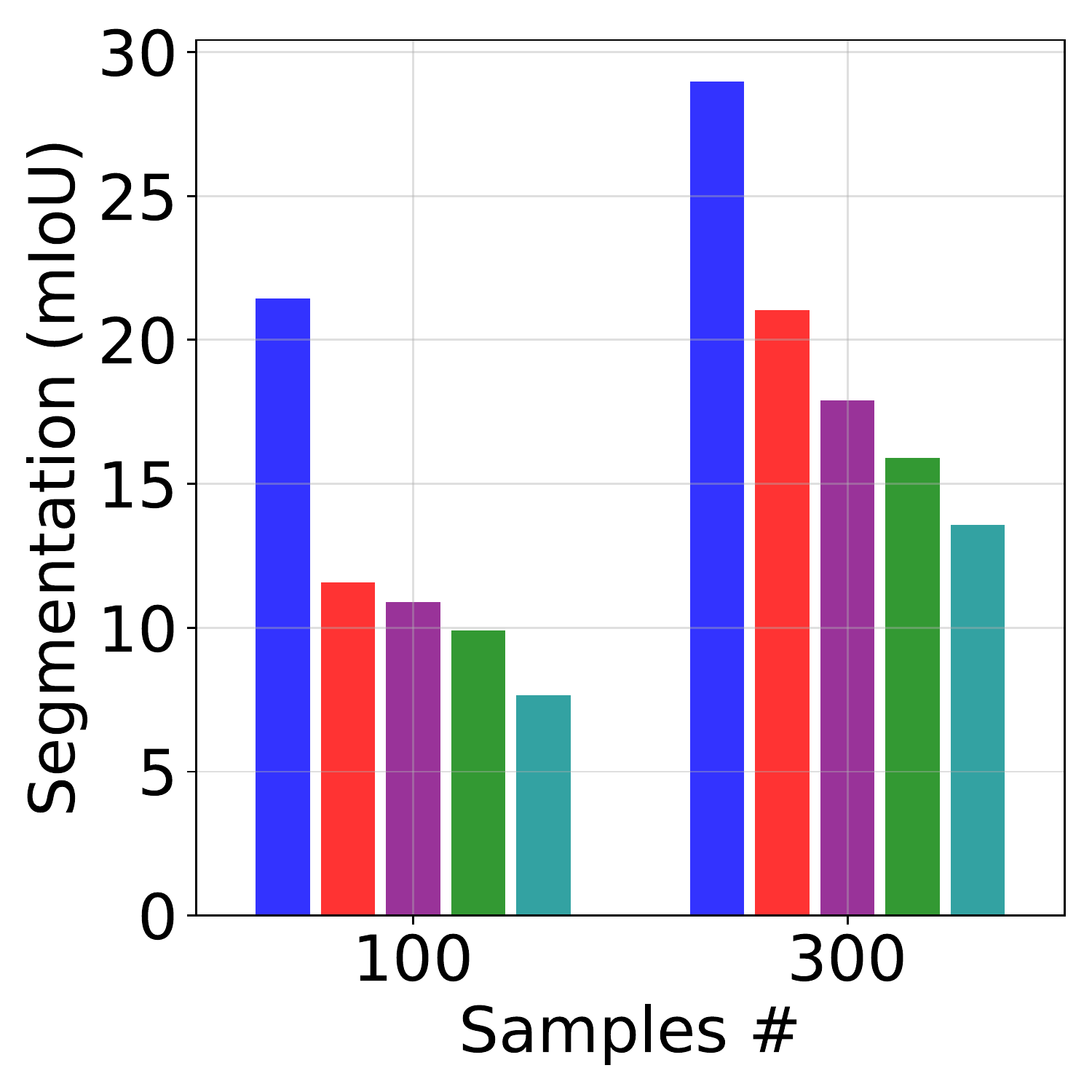} &
      \includegraphics[width=0.24\textwidth]{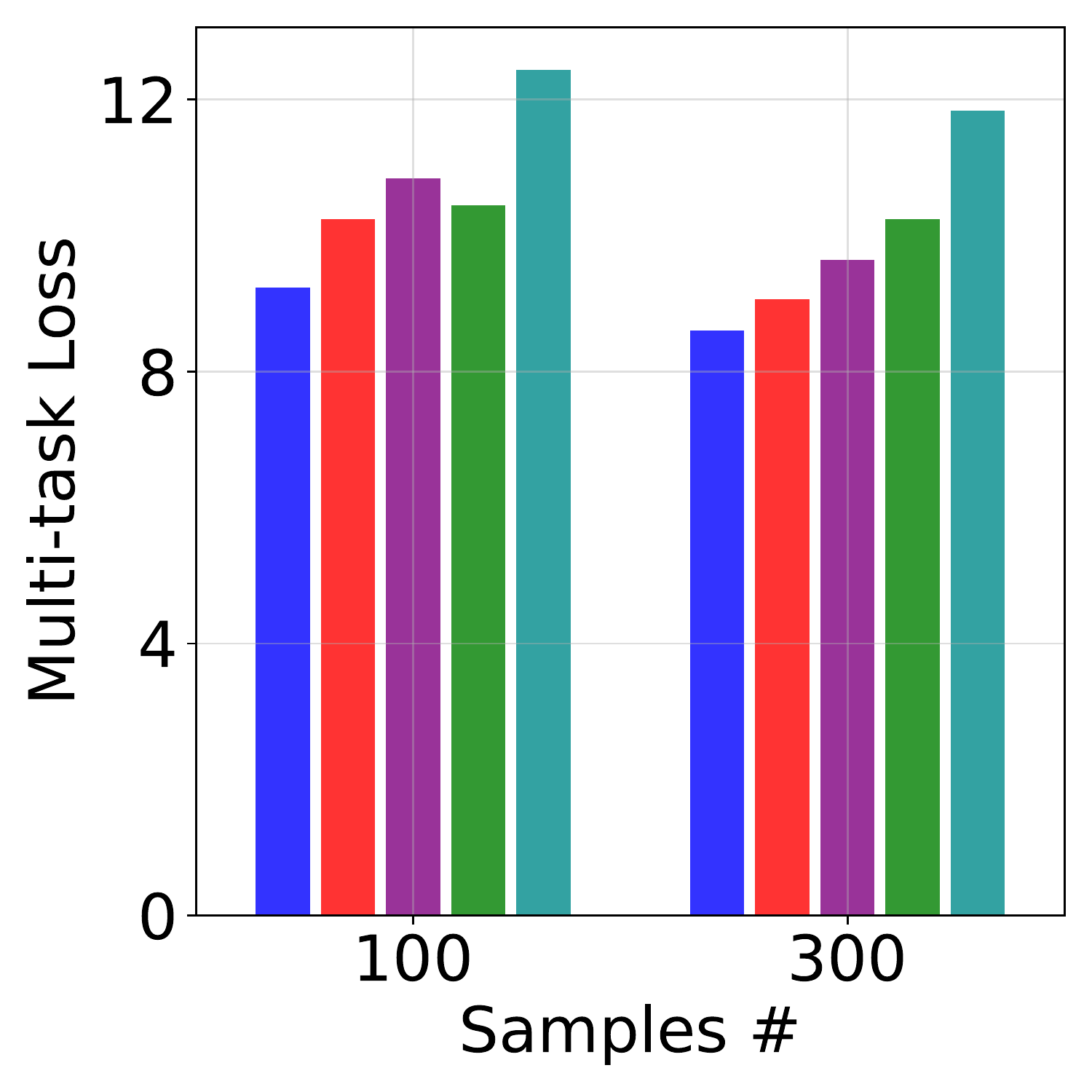}\\
    \end{tabular}
    \vspace{-5mm}
    \caption{
    \textbf{Picking informative samples.} Depth, normal, and segmentation metrics computed for the samples the various methods pick after the network has been trained using either 100 or 300 fully labelled images. On average \pale{} picks samples and modalities for which the metrics are worse, which makes labelling them particularly desirable.}
    \label{fig:hard-samples-bars}
\end{figure}

\parag{Finding the Hardest Examples.} 

To show that our approach picks the most difficult samples during the AL step, we use again the NYUD-v2 dataset of Section~\ref{sec:nyud}. As before, we train the network using either 100 or 300 of samples with all modalities labelled. Using either \pale{} or one of the baselines, we then pick 100 additional samples, that is 300 labels to be provided. In Fig.~\ref{fig:hard-samples-bars}, we plot how far off the predictions for these samples were. On average, our \pale{} picks samples with a greater depth and normal error and lower segmentation mIOU. In other words, in both cases, \pale{} finds samples and modalities on which the network performed worse and whose labelling is therefore most beneficial. 

\begin{figure}
    \includegraphics[width=0.47\textwidth]{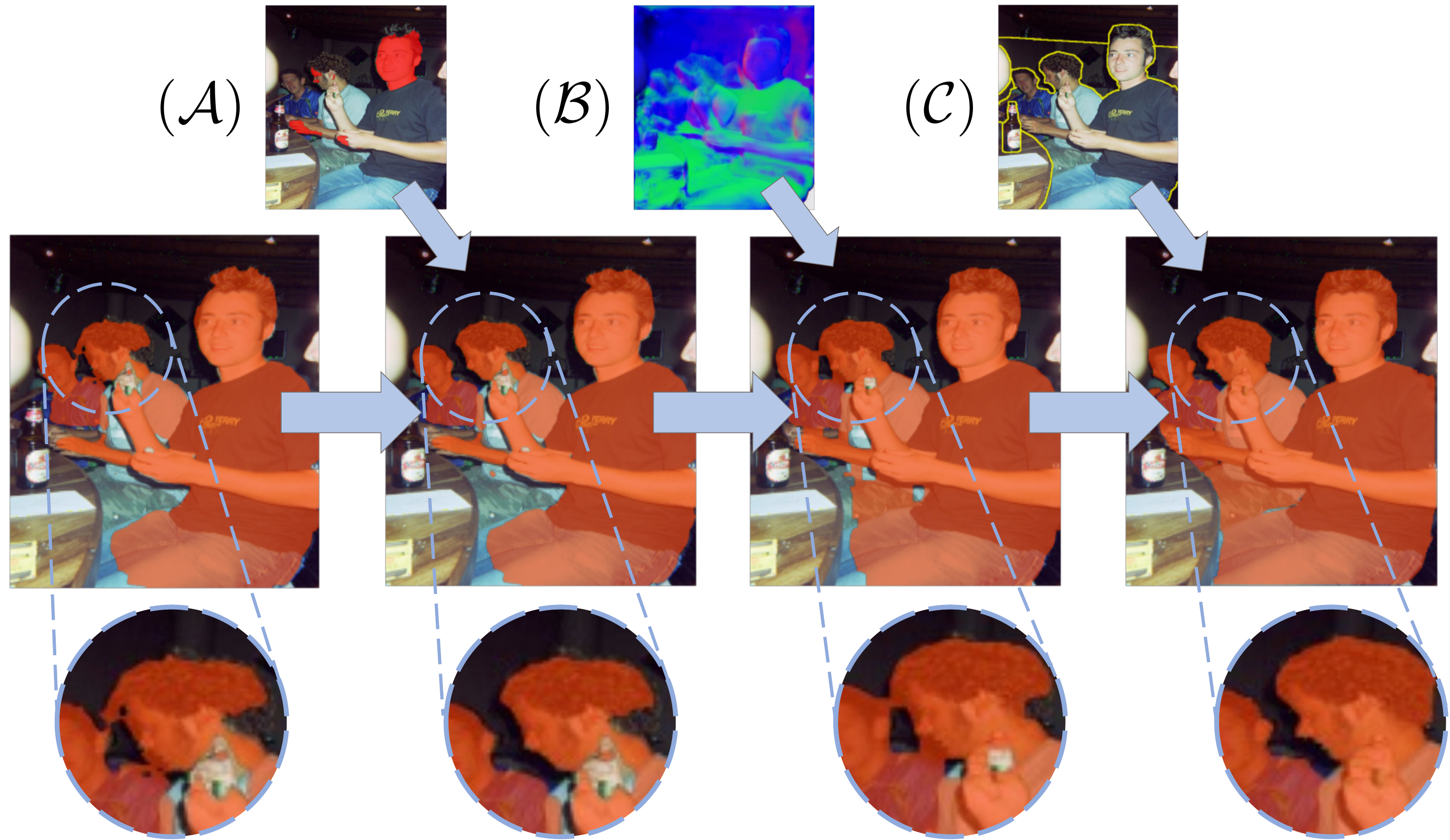}
    \centering
    \caption{
    \textbf{Improving segmentation quality.} Starting with an initial PAD-Net prediction (left), we gradually provide additional labels as inputs to the network: ($\mathcal{A}$) saliency, ($\mathcal{B}$) normals, and ($\mathcal{C}$) edges. As modalities are added, the segmentation quality improves. 
    }
    \label{fig:segm improvement}
\end{figure}

\begin{figure}
\centering
\begin{tabular}{@{}c@{}c@{}}
  \includegraphics[width=0.24\textwidth]{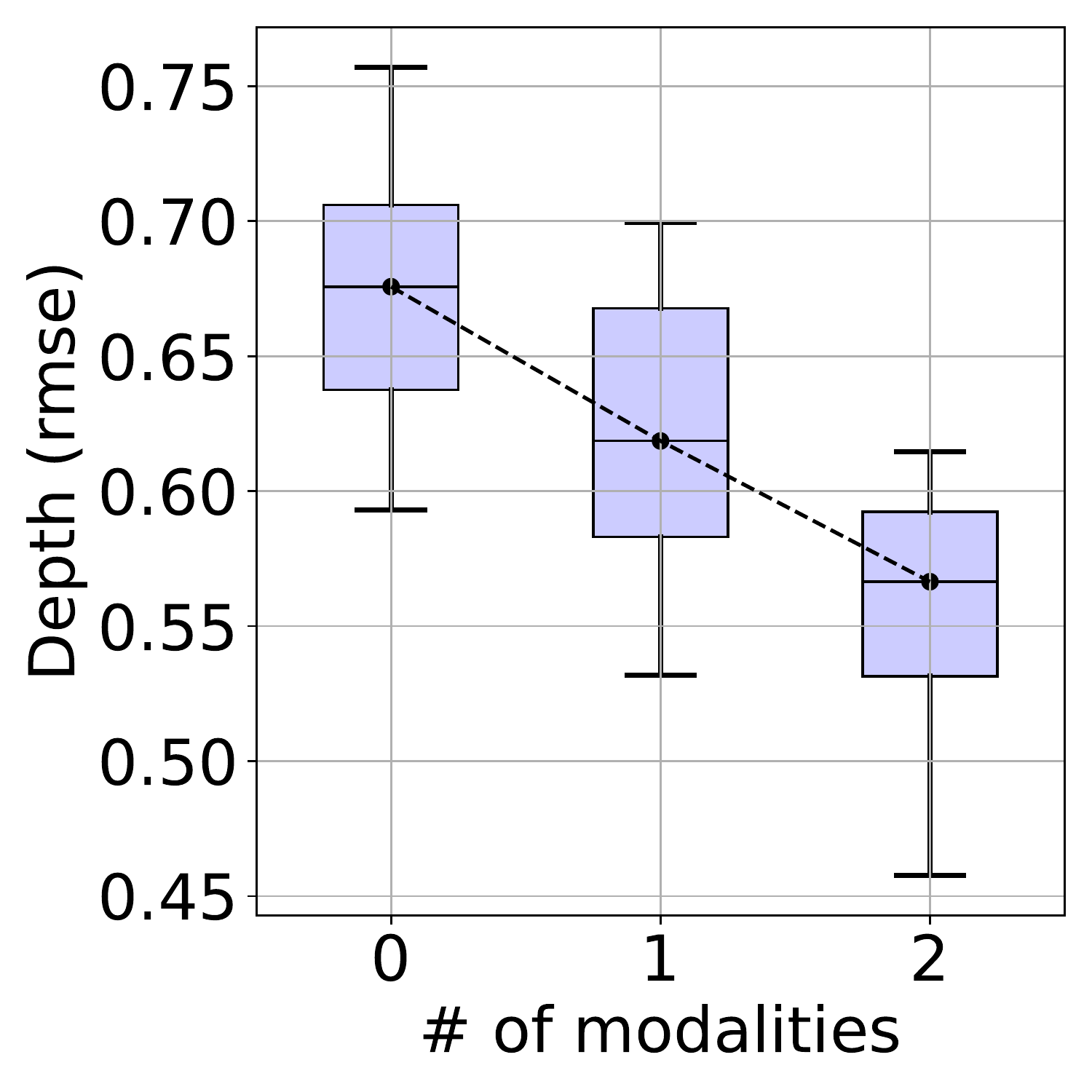} &    \includegraphics[width=0.24\textwidth]{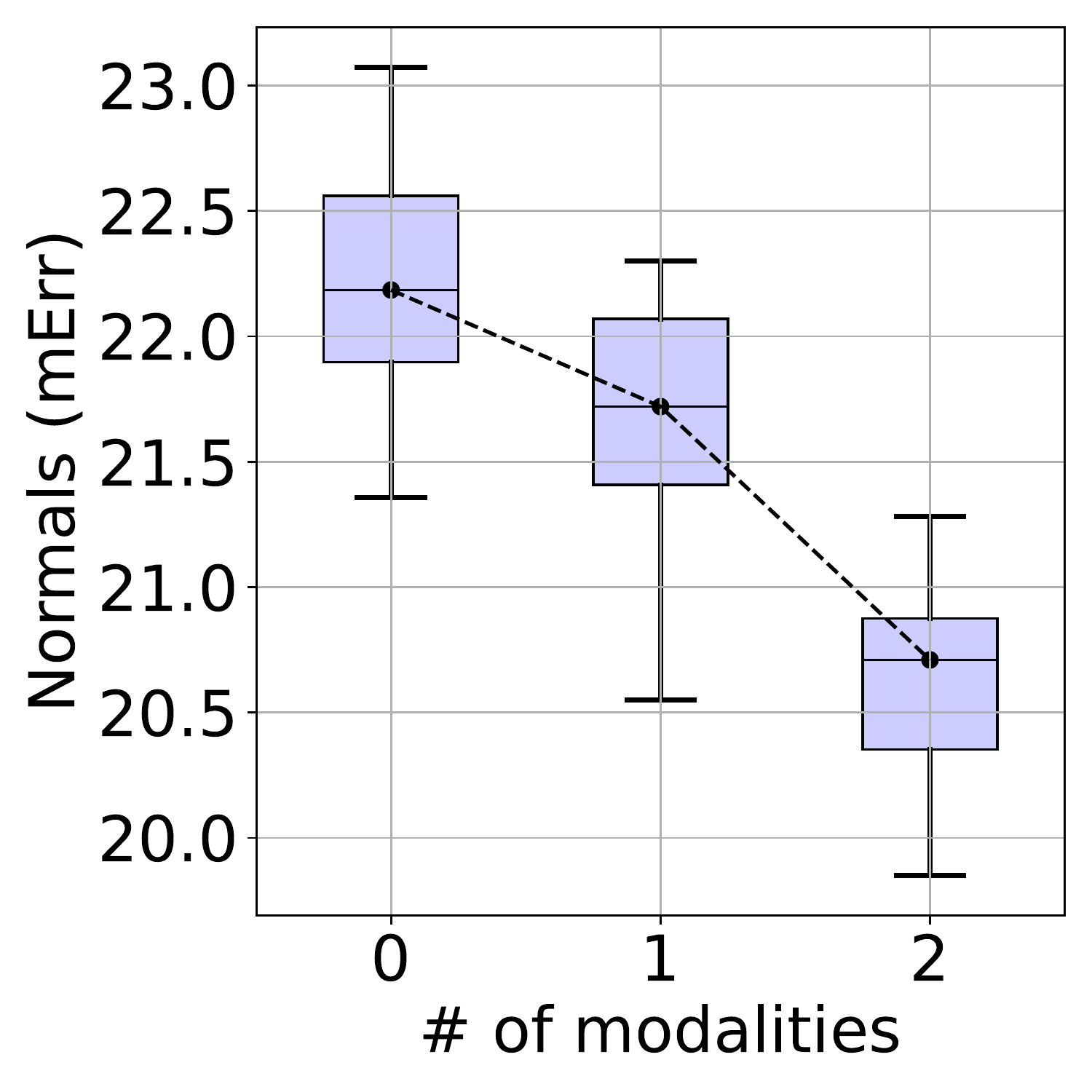}\\
  \includegraphics[width=0.24\textwidth]{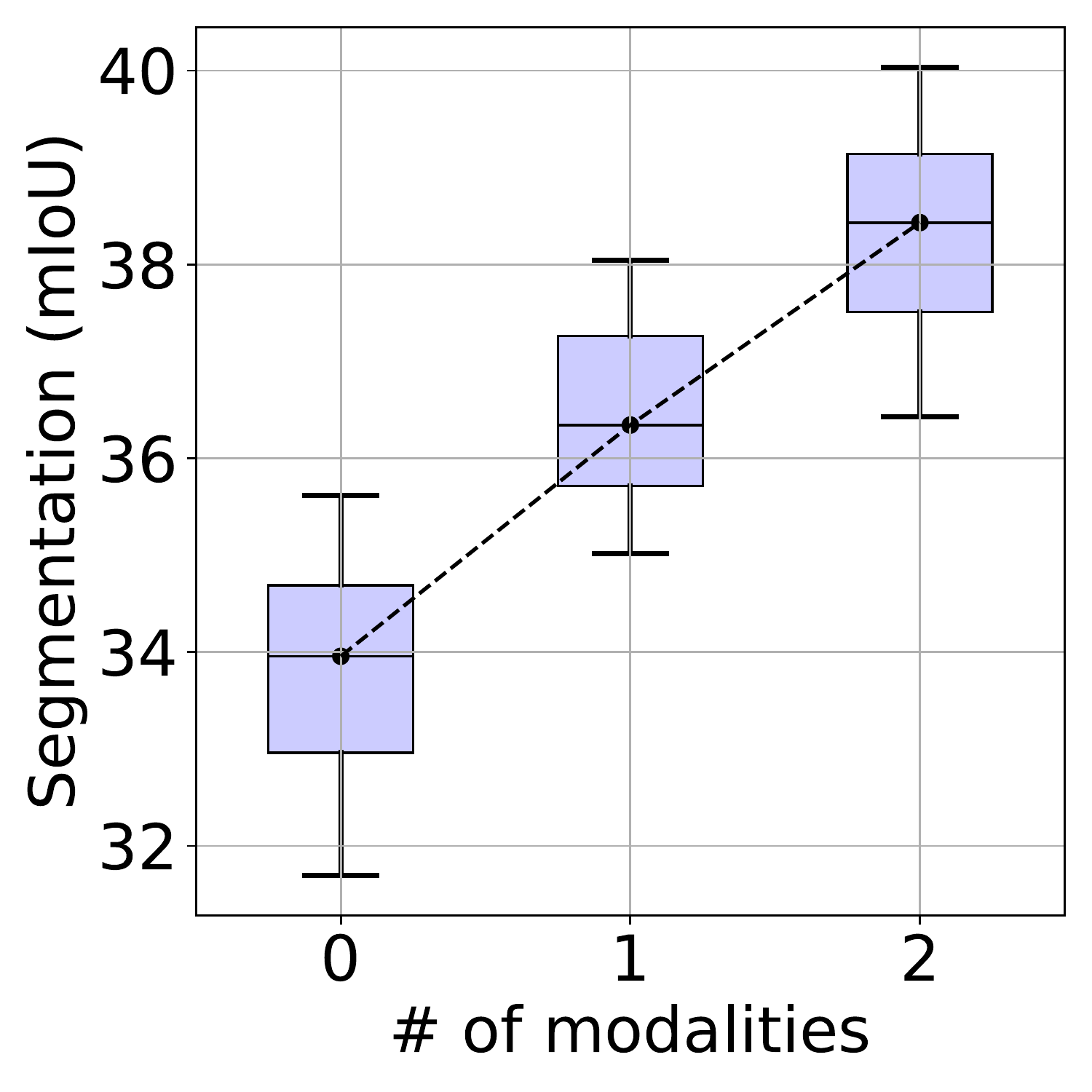} &    \includegraphics[width=0.24\textwidth]{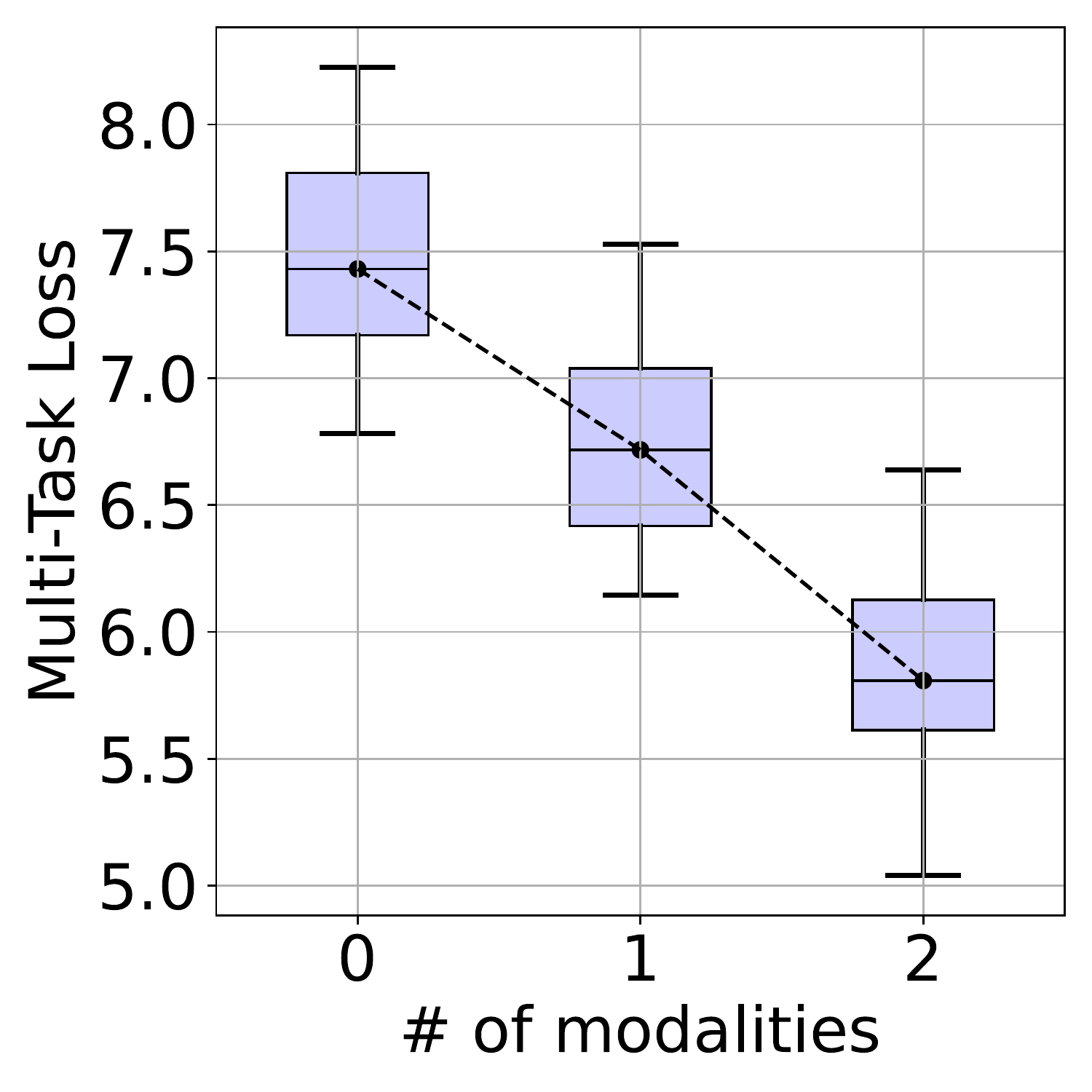}\\
\end{tabular}
\vspace{-5mm}
\caption{
\textbf{Improving inference by providing partial labels.} We train our model using all of the training data in NYUD-v2. The blue lines represent the metrics calculated on the evaluation set. The shaded areas depict the corresponding standard deviations over samples. x-axis represents the number of labelled modalities used during inference. These plots demonstrate that adding targets as inputs helps infer other modalities more accurately. }
\label{fig:loss_vs_labels}
\end{figure}

\parag{Improved Inference.}

As discussed in Section~\ref{sec:partial_inference}, partial labels can be used to make predictions more accurate at inference time, as shown in Fig.~\ref{fig:segm improvement}. To quantify this, we retrain our model using all training data available in $\bL$ and $\bU$ for NYUDv2. On the test set, we then provide different combinations of labels along with the input image to evaluate their ability to improve the prediction. More formally, during inference, we provide the input image $\bx_{i}$ and a subset of labels $\{\by^{t_{1}}_{i}, \by^{t_{2}}_{i}, ... , \by^{t_{M}}_{i}\}$ to be used to infer the other modalities. Intuitively, the more modalities we provide during inference, the more accurate the model's outputs should be. To support this claim quantitatively, we iterate over all available subsets of sizes from $0$ to $K-1$, use them to predict other labels, and calculate metrics for each. We plot the results in Fig.~\ref{fig:loss_vs_labels}. They clearly show that the more modalities we provide as input, the better the result.


\begin{figure}
\centering
\begin{tabular}{@{}c@{}c@{}}
  \includegraphics[width=0.24\textwidth]{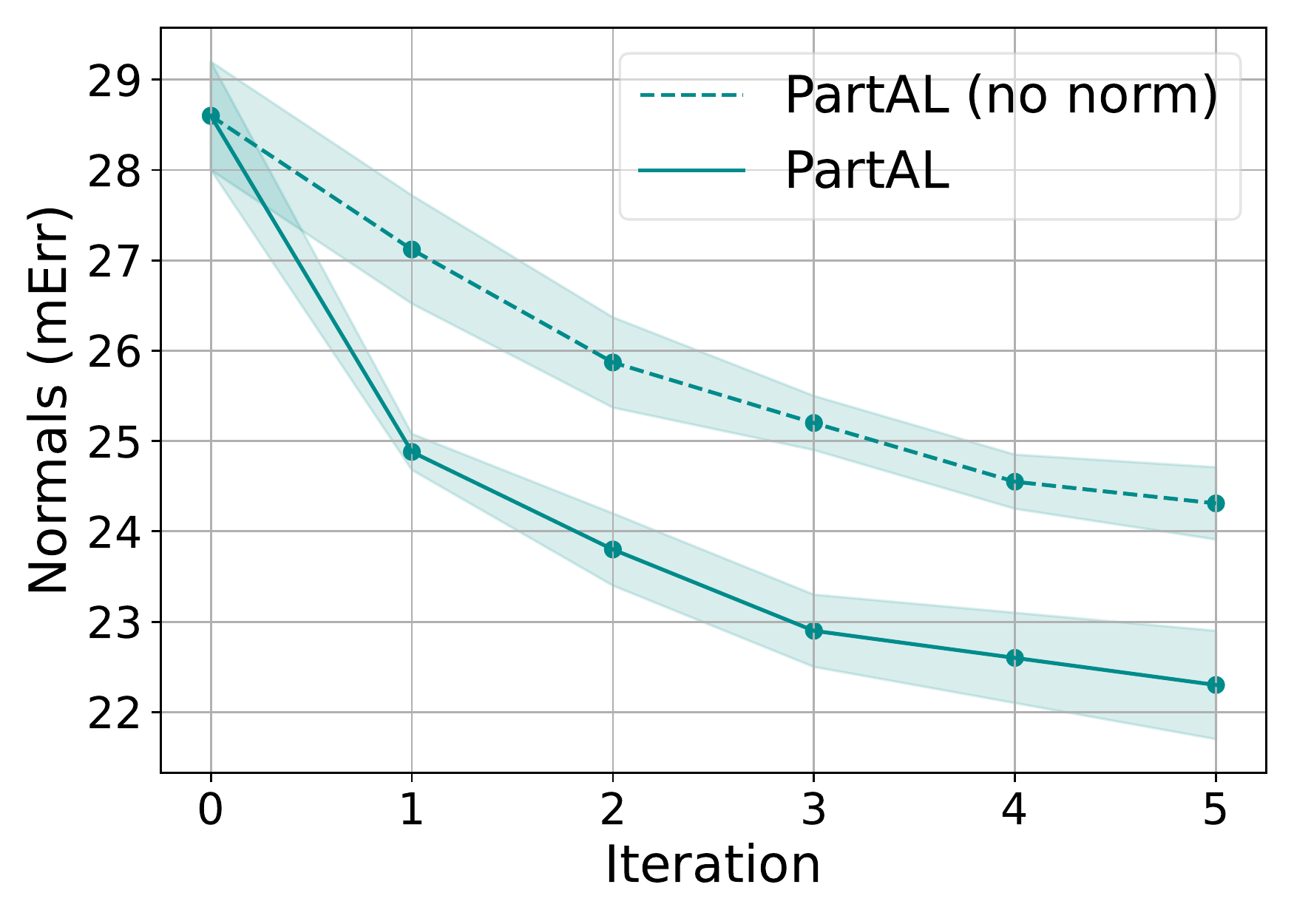} &    \includegraphics[width=0.24\textwidth]{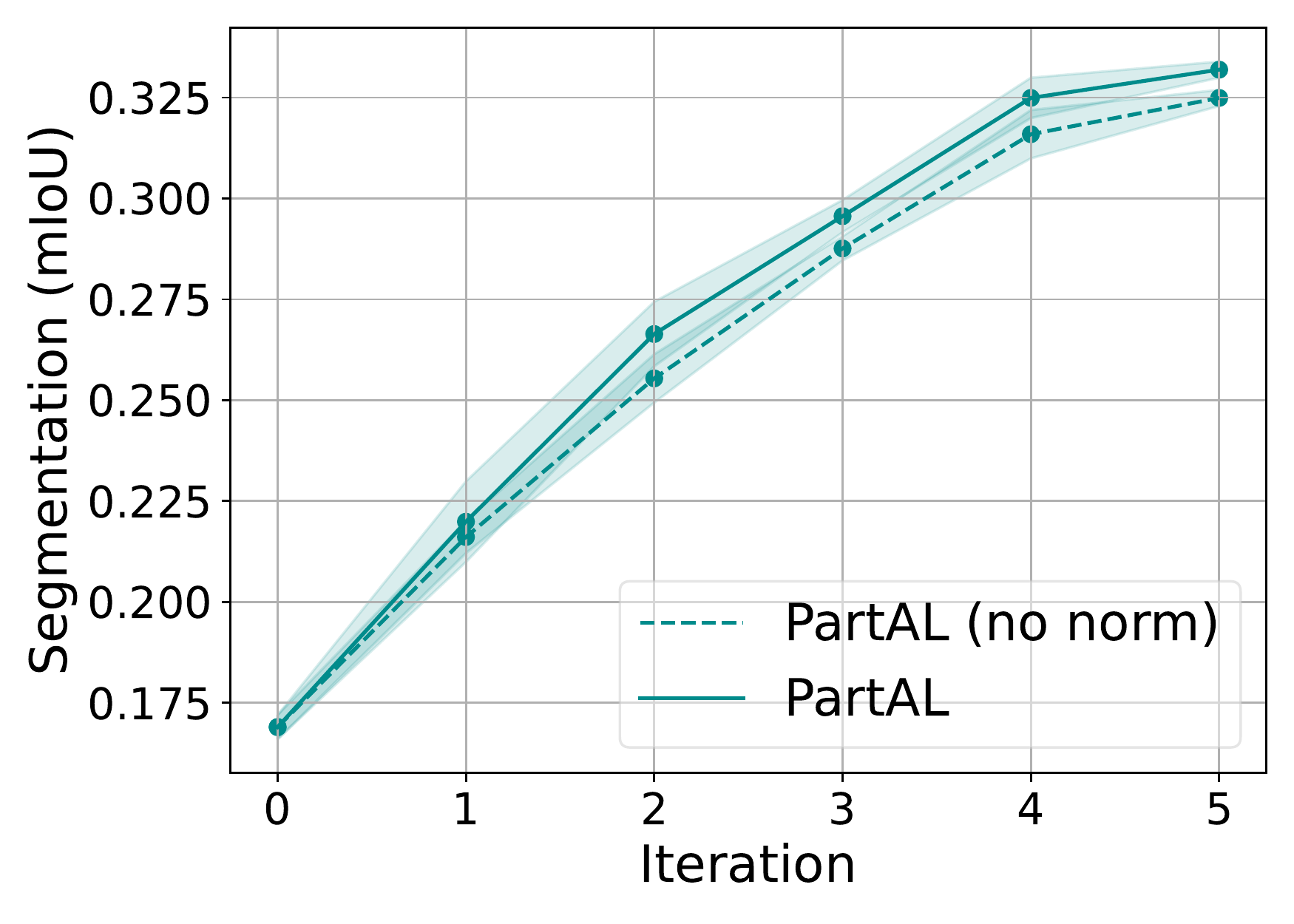}\\[1pt]
  \includegraphics[width=0.24\textwidth]{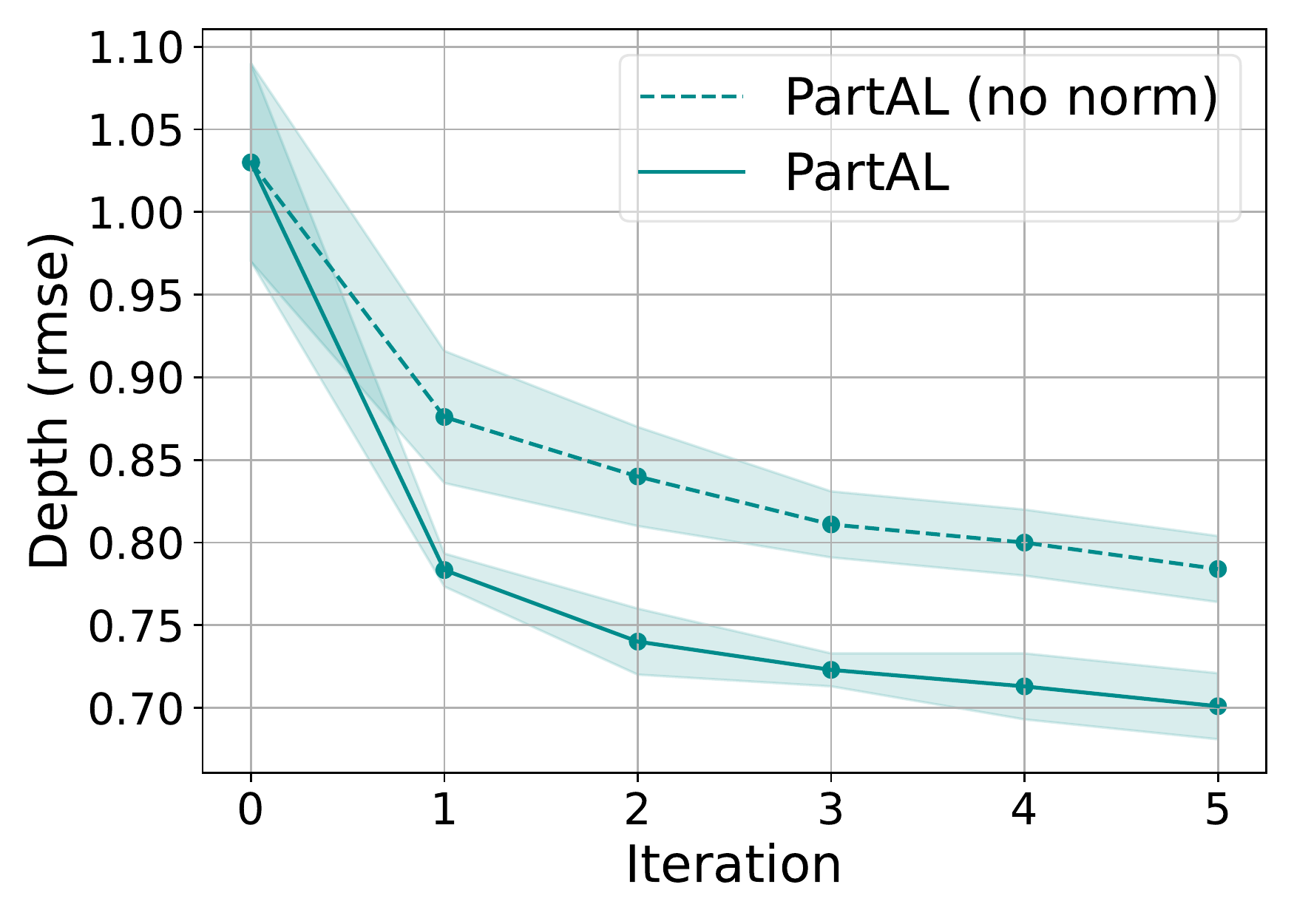} &    \includegraphics[width=0.24\textwidth]{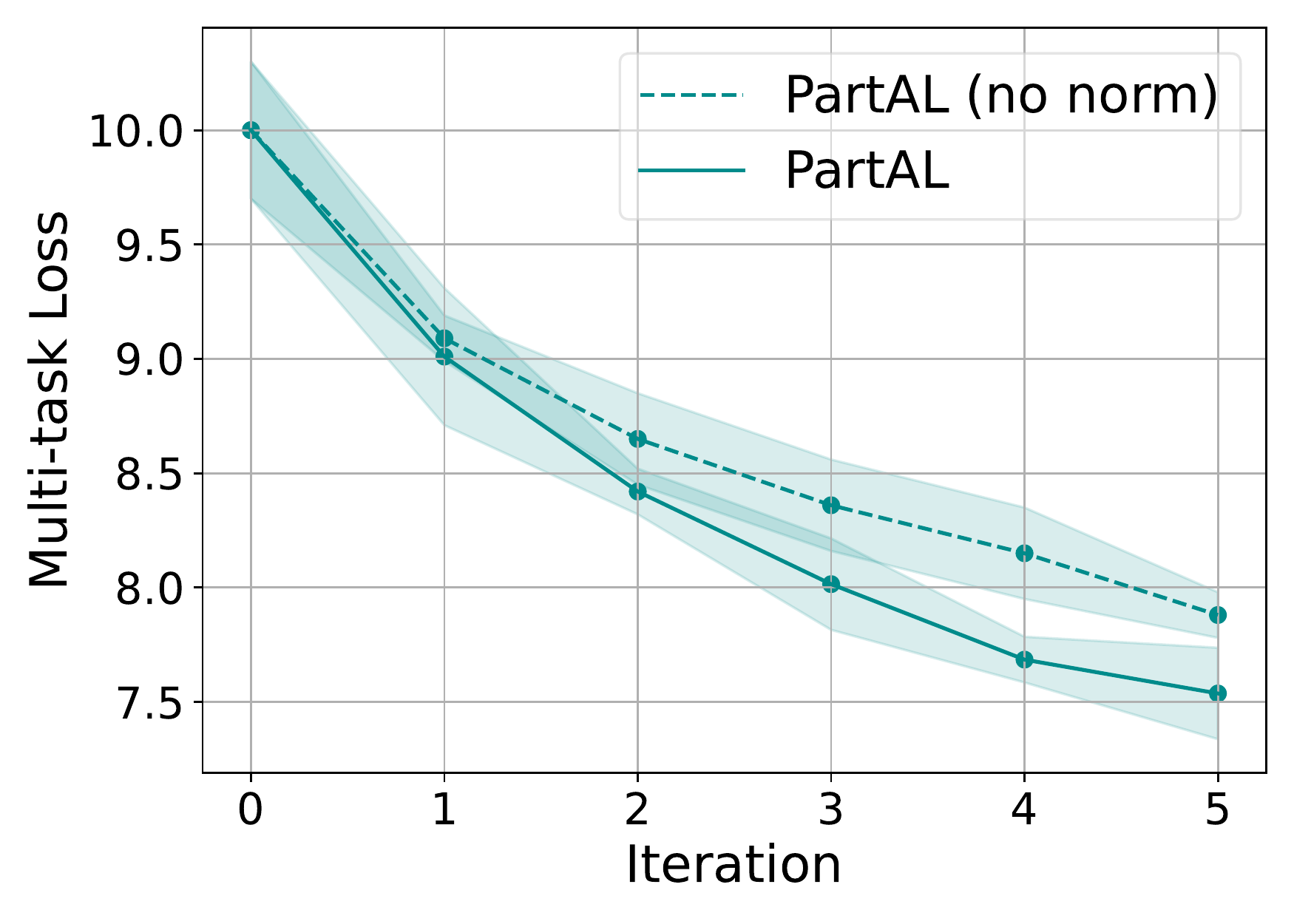}\\[1pt]
\end{tabular}
\vspace{-5mm}
\caption{
\textbf{Uncertainty normalization.} Removing uncertainty normalization from \pale{} significantly reduces performance. }
\label{fig:unc-norm}
\end{figure}

\parag{Uncertainty Normalization.}

As mentioned in Section~\ref{sec:choosing}, uncertainties from different modalities are not directly comparable in general and we have to normalize them to make them comparable. To demonstrate the importance of this,  we re-ran the experiments of Section~\ref{sec:nyud}, but without normalization. We plot the results with and without normalization in Fig.~\ref{fig:unc-norm}. Unsurprisingly the latter are substantially worse, which shows the importance of our normalization mechanism.  


\begin{figure}
\centering
\begin{tabular}{@{}c@{}c@{}}
  \includegraphics[width=0.24\textwidth]{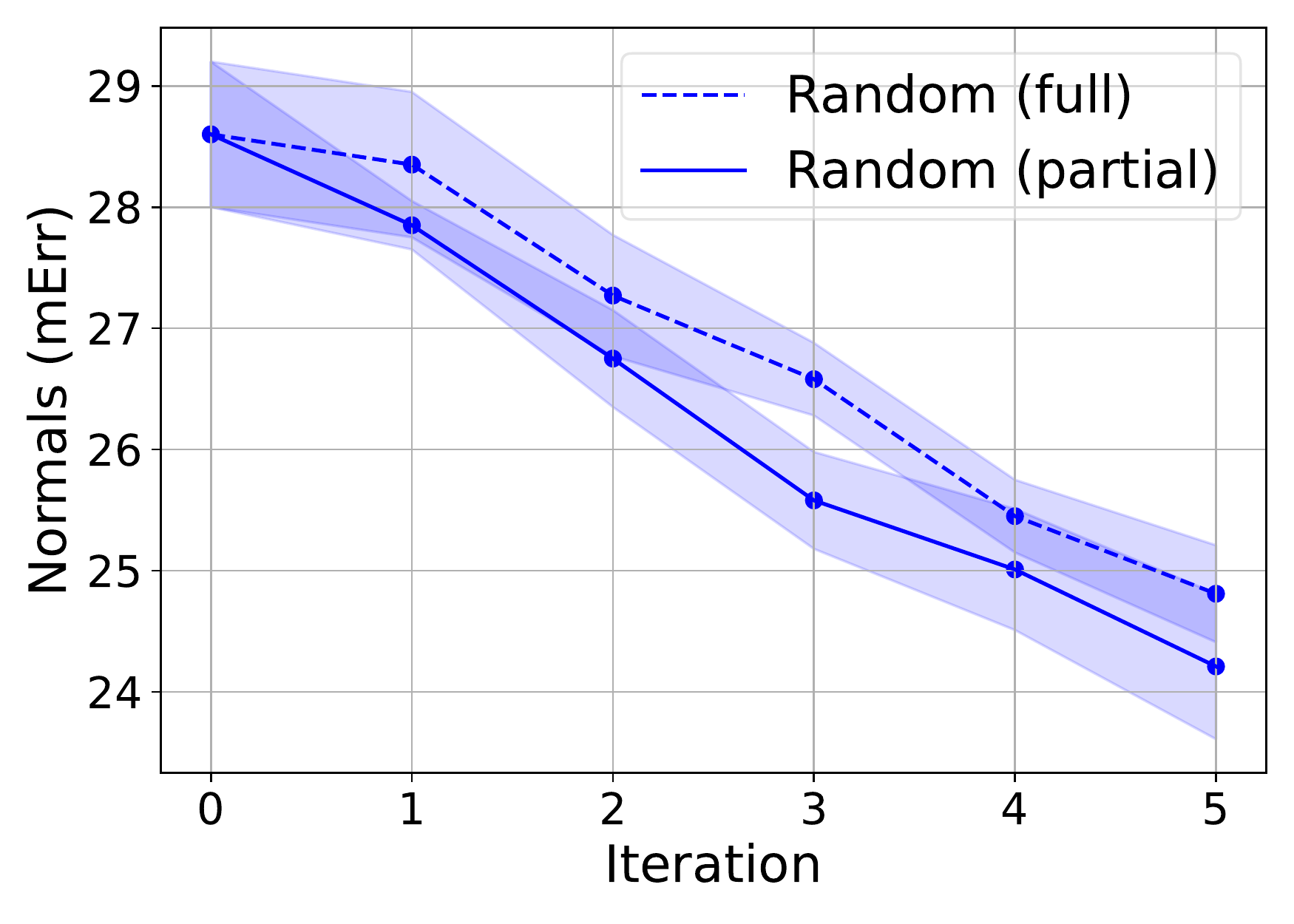} &    
  \includegraphics[width=0.24\textwidth]{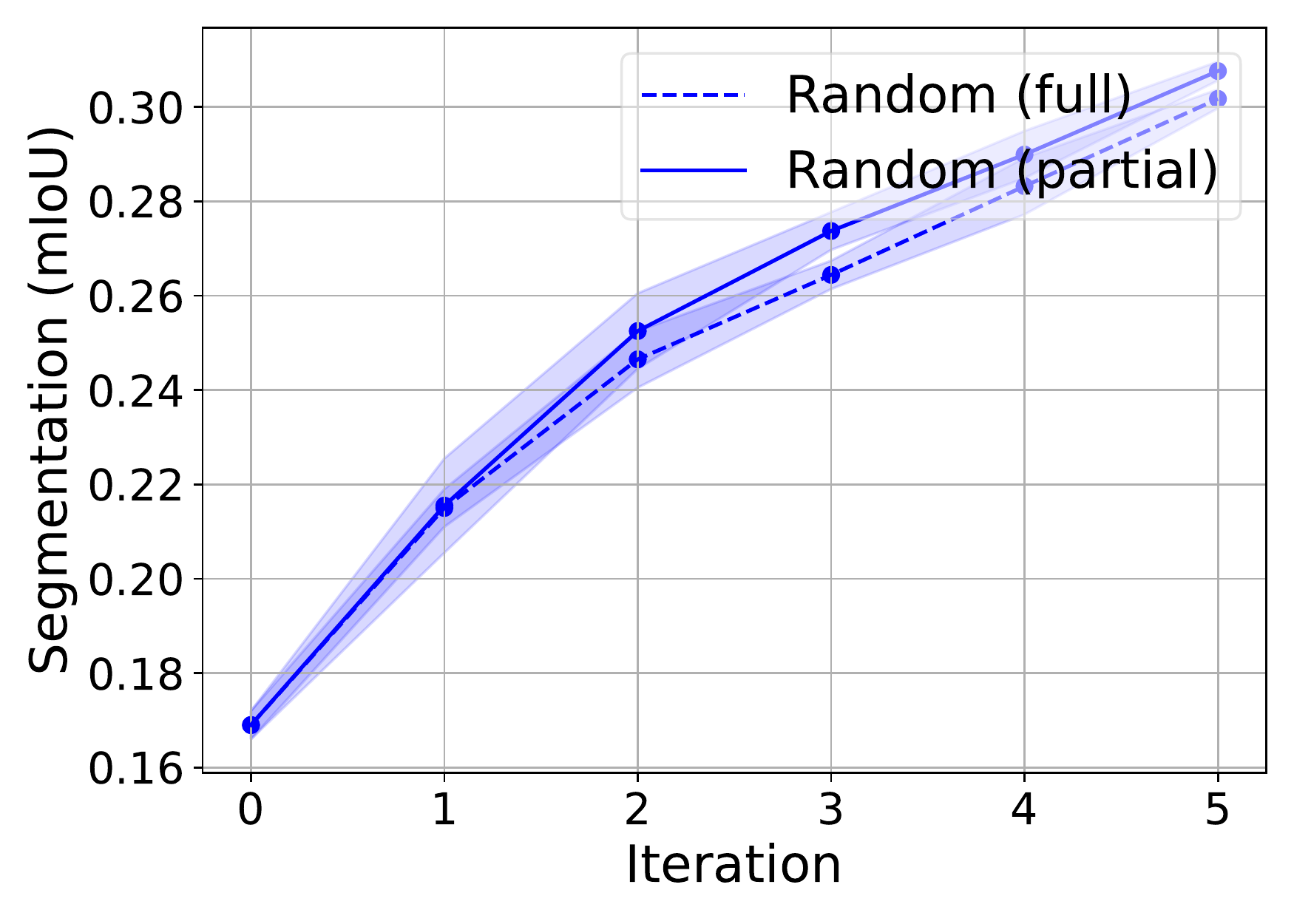}\\[-3mm]
  \includegraphics[width=0.24\textwidth]{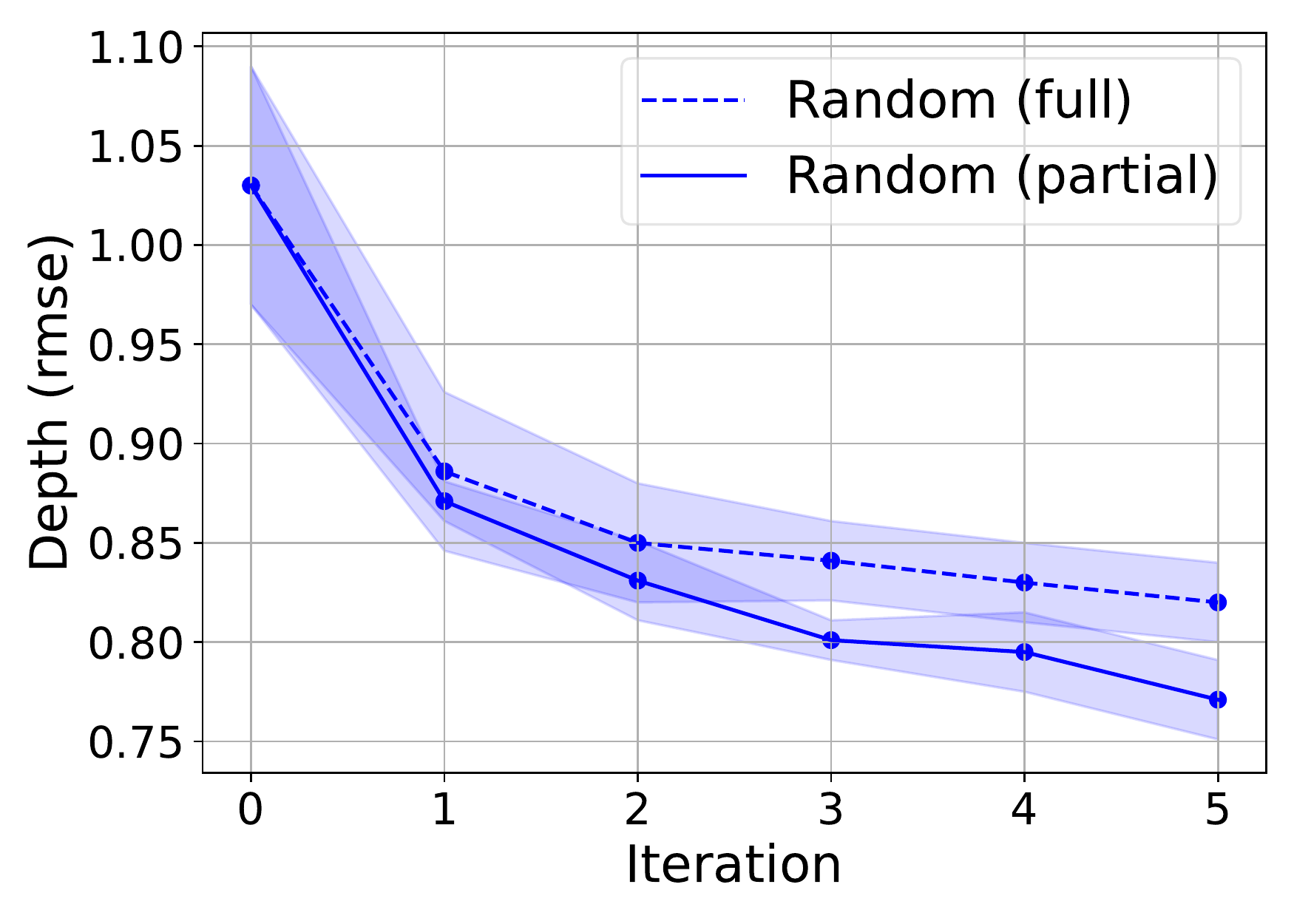} &    
  \includegraphics[width=0.24\textwidth]{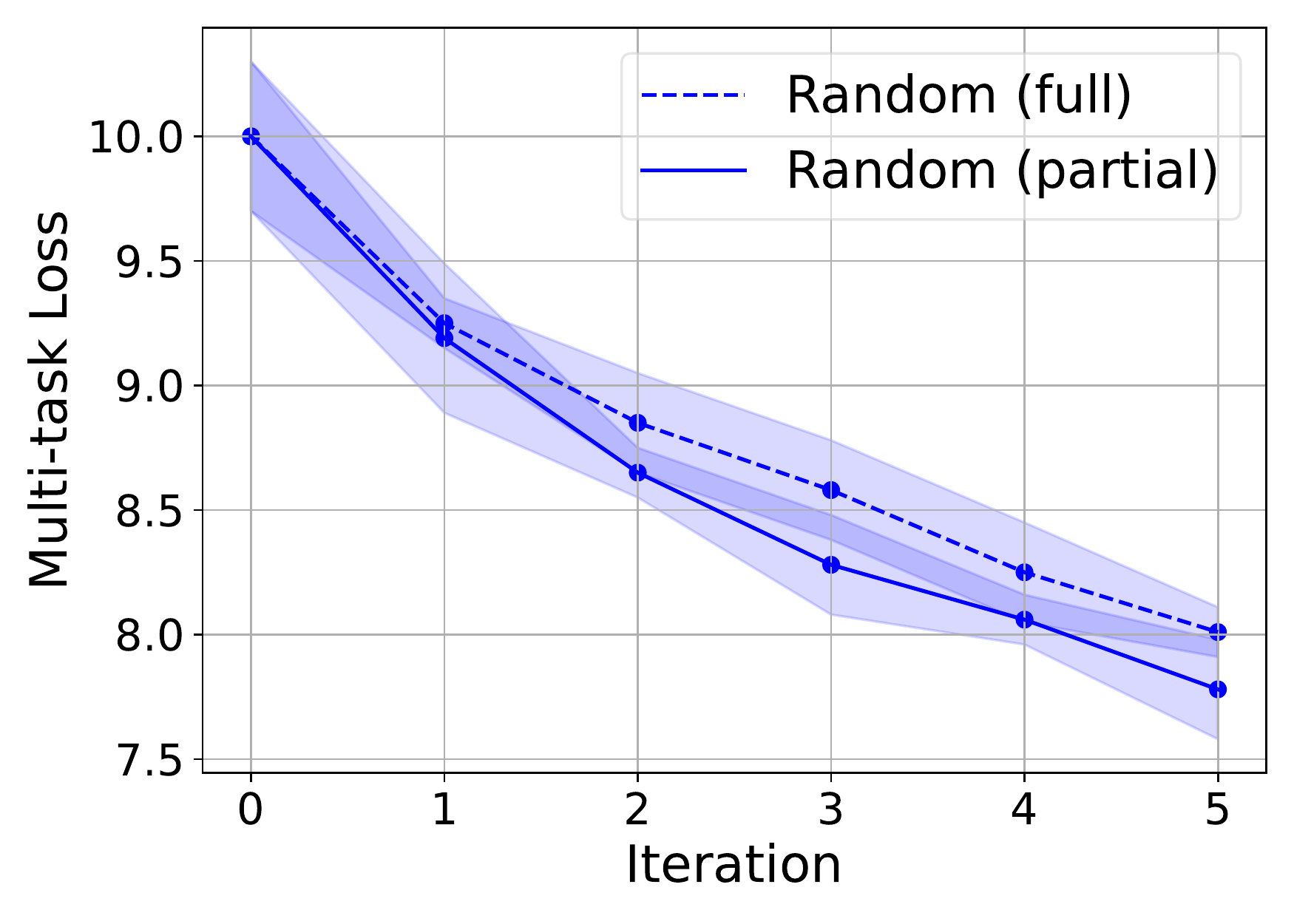}\\
\end{tabular}
\vspace{-6mm}
\caption{
\textbf{Partial labels only.} Using partial labels delivers an improvement event without taking uncertainty into account. 
}
\vspace{-2mm}
\label{fig:random-partial}
\end{figure}

\parag{Partial Labels Improve Active Learning.}

A strength of our approach is its ability to exploiting correlations between modalities. To demonstrate this by itself, we re-ran the experiments of Section~\ref{sec:nyud} with the \textit{Random} baseline modified to use partial labels. In other words, we use our pipeline but instead of picking modalities based on uncertainty we chose them randomly.  As shown in Fig.~\ref{fig:random-partial} demonstrates, this already provides an improvement. 


\section{Conclusion}

We have proposed  \pale{}, a method for multi-task active learning that uses partial labelling as an alternative to current approaches. While being easy-to-implement and not requiring any additional computational overhead, \pale{} demonstrates state-of-the-art results in terms of data efficiency and model accuracy on several datasets and tasks. 

A key component of \pale{} is the uncertainty estimator that we use to decide which samples and modalities to pick. In our current implementation, we rely on MC-Dropout~\cite{Gal16a}, which only imposes a computational burden that is relatively small compared to a more involved but more reliable technique such as Ensembles~\cite{Lakshminarayanan17}. There have been recent attempts at increasing the reliability of MC-Dropout~\cite{Durasov21a,Wen2020a} and we will investigate the use of such techniques to further boost the performance of \pale{}. Apart from ensembling-based techniques, a number of sampling-free approaches were recently proposed for deep learning models~\cite{Liu20,Tagasovska19,Postels19}  which are potentially able to reduce computational burdens of AL. Despite a variety of uncertainty estimation methods, none of them is specifically designed for multi-tasking, which is an important and promising direction in the development of multi-task active learning techniques.



\clearpage

{\small
\bibliographystyle{ieee_fullname}
\bibliography{bib/string,bib/vision,bib/learning,bib/robotics,bib/stats,bib/cfd,bib/misc}

\begin{thebibliography}{10}\itemsep=-1pt

\bibitem{Beluch18}
W.H. Beluch, T. Genewein, A. N\"{u}rnberger, and J.M. K\"{o}hler.
\newblock {The Power of Ensembles for Active Learning in Image Classification}.
\newblock In {\em Conference on Computer Vision and Pattern Recognition}, 2018.

\bibitem{Caruana97}
R. Caruana.
\newblock {Multitask Learning}.
\newblock {\em Machine Learning}, 28, 1997.

\bibitem{Chen14a}
X. Chen, R. Mottaghi, X. Liu, S. Fidler, R. Urtasun, and A. Yuille.
\newblock {Detect What You Can: Detecting and Representing Objects Using
  Holistic Models and Body Parts}.
\newblock In {\em Conference on Computer Vision and Pattern Recognition}, 2014.

\bibitem{Dor20}
L.~E. Dor, A. Halfon, A. Gera, E. Shnarch, L. Dankin, L. Choshen, M.
  Danilevsky, R. Aharonov, Y. Katz, and N. Slonim.
\newblock {Active Learning for BERT: An Empirical Study}.
\newblock In {\em Conference on Empirical Methods in Natural Language
  Processing}, 2020.

\bibitem{Duong15}
L. Duong, T.Cohn, S. Bird, and P. Cook.
\newblock {Low Resource Dependency Parsing: Cross-Lingual Parameter Sharing in
  a Neural Network Parser}.
\newblock In {\em International Joint Conference on Natural Language
  Processing}, 2015.

\bibitem{Durasov21a}
N. Durasov, T. Bagautdinov, P. Baque, and P. Fua.
\newblock {Masksembles for Uncertainty Estimation}.
\newblock In {\em Conference on Computer Vision and Pattern Recognition}, 2021.

\bibitem{Durasov21b}
N. Durasov, A. Lukoyanov, J. Donier, and P. Fua.
\newblock {DEBOSH: Deep Bayesian Shape Optimization}.
\newblock In {\em arXiv Preprint}, 2021.

\bibitem{Everingham10}
Mark Everingham, Luc~Van Gool, Christopher K.~I. Williams, John~M. Winn, and
  Andrew Zisserman.
\newblock {The Pascal Visual Object Classes (VOC) Challenge}.
\newblock {\em International Journal of Computer Vision}, 88(2):303--338, 2010.

\bibitem{Flores20}
R.~A. Flores, C. Paolucci, K.~T. Winther, A. Jain, J.~A.~G. Torres, M. Aykol,
  J. Montoya, J.~K. N{\o}rskov, M. Bajdich, and T. Bligaard.
\newblock {Active Learning Accelerated Discovery of Stable Iridium Oxide
  Polymorphs for the Oxygen Evolution Reaction}.
\newblock {\em Chemistry of Materials}, 2020.

\bibitem{Gal16a}
Y. Gal and Z. Ghahramani.
\newblock {Dropout as a Bayesian Approximation: Representing Model Uncertainty
  in Deep Learning}.
\newblock In {\em International Conference on Machine Learning}, pages
  1050--1059, 2016.

\bibitem{Gal17}
Y. Gal, R. Islam, and Z. Ghahramani.
\newblock {Deep Bayesian Active Learning with Image Data}.
\newblock In {\em International Conference on Machine Learning}, pages
  1183--1192, 2017.

\bibitem{Gao19c}
Y. Gao, J. Ma, M. Zhao, W. Liu, and A.L. Yuille.
\newblock {Nddr-Cnn: Layerwise Feature Fusing in Multi-Task CNNs by Neural
  Discriminative Dimensionality Reduction}.
\newblock In {\em Conference on Computer Vision and Pattern Recognition}, 2019.

\bibitem{Hao20b}
Z. Hao, C. Lu, Z. Huang, H. Wang, Z. Hu, Q. Liu, E. Chen, and C. Lee.
\newblock {ASGN: An Active Semi-Supervised Graph Neural Network for Molecular
  Property Prediction}.
\newblock In {\em International Conference on Knowledge Discovery and Data
  Mining}, 2020.

\bibitem{Kaushal19}
V. Kaushal, R. Iyer, S. Kothawade, R. Mahadev, K. Doctor, and G. Ramakrishnan.
\newblock {Learning from Less Data: A Unified Data Subset Selection and Active
  Learning Framework for Computer Vision}.
\newblock In {\em IEEE Winter Conference on Applications of Computer Vision},
  2019.

\bibitem{Kendall18}
A. Kendall, Y. Gal, and R. Cipolla.
\newblock {Multi-Task Learning Using Uncertainty to Weigh Losses for Scene
  Geometry and Semantics}.
\newblock In {\em Conference on Computer Vision and Pattern Recognition}, 2018.

\bibitem{Kokkinos17}
I. Kokkinos.
\newblock {Ubernet: Training a Universal Convolutional Neural Network for Low-,
  Mid-, and High-Level Vision Using Diverse Datasets and Limited Memory}.
\newblock In {\em Conference on Computer Vision and Pattern Recognition}, 2017.

\bibitem{Kumar86}
Uma Kumar, Vinod Kumar, and J~N Kapur.
\newblock {Normalized Measures of Entropy}.
\newblock {\em International Journal Of General System}, 12(1):55--69, 1986.

\bibitem{Lakshminarayanan17}
B. Lakshminarayanan, A. Pritzel, and C. Blundell.
\newblock {Simple and Scalable Predictive Uncertainty Estimation Using Deep
  Ensembles}.
\newblock In {\em Advances in Neural Information Processing Systems}, 2017.

\bibitem{LeCam12}
Lucien Le~Cam.
\newblock {\em Asymptotic Methods in Statistical Decision Theory}.
\newblock Springer Science \& Business Media, 2012.

\bibitem{Lewis94}
D.D. Lewis, D. D., and W.A. Gale.
\newblock {A Sequential Algorithm for Training Text Classifiers}.
\newblock In {\em ACM SIGIR proceedings on Research and Development in
  Information Retrieval}, 1994.

\bibitem{Li13f}
X. Li and Y. Guo.
\newblock {Adaptive Active Learning for Image Classification}.
\newblock In {\em Conference on Computer Vision and Pattern Recognition}, 2013.

\bibitem{Liu20}
J. Liu, Z. Lin, S. Padhy, D. Tran, T.~B. Weiss, and B. Lakshminarayanan.
\newblock {Simple and Principled Uncertainty Estimation with Deterministic Deep
  Learning via Distance Awareness}.
\newblock In {\em Advances in Neural Information Processing Systems}, 2020.

\bibitem{MacKay95}
D.~J. Mackay.
\newblock {Bayesian Neural Networks and Density Networks}.
\newblock {\em {Nuclear Instruments and Methods in Physics Research Section A:
  Accelerators, Spectrometers, Detectors and Associated Equipment}},
  354(1):73--80, 1995.

\bibitem{Martin04}
D. Martin, C. Fowlkes, and J. Malik.
\newblock {Learning to Detect Natural Image Boundaries Using Local Brightness,
  Color and Texture Cues}.
\newblock {\em IEEE Transactions on Pattern Analysis and Machine Intelligence},
  26(5), 2004.

\bibitem{Misra16a}
I. Misra, A. Shrivastava, A. Gupta, and M. Hebert.
\newblock {Cross-Stitch Networks for Multi-Task Learning}.
\newblock In {\em Conference on Computer Vision and Pattern Recognition}, pages
  3994--4003, 2016.

\bibitem{Postels19}
J. Postels, F. Ferroni, H. Coskun, N. Navab, and F. Tombari.
\newblock {Sampling-Free Epistemic Uncertainty Estimation Using Approximated
  Variance Propagation}.
\newblock In {\em Conference on Computer Vision and Pattern Recognition}, pages
  2931--2940, 2019.

\bibitem{Qi08}
G.-J. Qi, X.-S. Hua, Y. Rui, J. Tang, and H.-J. Zhang.
\newblock {Two-Dimensional Active Learning for Image Classification}.
\newblock In {\em Conference on Computer Vision and Pattern Recognition}, 2008.

\bibitem{Reichart08}
R. Reichart, K. Tomanek, U. Hahn, and A. Rappoport.
\newblock {Multi-Task Active Learning for Linguistic Annotations}.
\newblock In {\em ACL}, 2008.

\bibitem{Ruder17}
S. Ruder.
\newblock {An Overview of Multi-Task Learning in Deep Neural Networks}.
\newblock In {\em arXiv Preprint}, 2017.

\bibitem{Sener17}
O. Sener and S. Savarese.
\newblock {Active Learning for Convolutional Neural Networks: A Core-Set
  Approach}.
\newblock In {\em arXiv Preprint}, 2017.

\bibitem{Settles09a}
B. Settles.
\newblock {Active Learning Literature Survey}.
\newblock Technical report, University of Wisconsin-Madison Department of
  Computer Sciences, 2009.

\bibitem{Settles08b}
B. Settles and M. Craven.
\newblock {An Analysis of Active Learning Strategies for Sequence Labeling
  Tasks}.
\newblock In {\em Conference on Empirical Methods in Natural Language
  Processing}, pages 1070--1079, 2008.

\bibitem{Shannon48}
Claude~Elwood Shannon.
\newblock {A mathematical theory of communication}.
\newblock {\em The Bell System Technical Journal}, 27(3):379--423, 1948.

\bibitem{Siddhant18}
A. Siddhant and Z.~C. Lipton.
\newblock {Deep Bayesian Active Learning for Natural Language Processing:
  Results of a Large-Scale Empirical Study}.
\newblock In {\em arXiv Preprint}, 2018.

\bibitem{Silberman12}
N. Silberman, D. Hoiem, P. Kohli, and R. Fergus.
\newblock {Indoor Segmentation and Support Inference from RGBD Images}.
\newblock In {\em European Conference on Computer Vision}, 2012.

\bibitem{Srivastava14}
N. Srivastava, G. Hinton, A. Krizhevsky, I. Sutskever, and R. Salakhutdinov.
\newblock {Dropout: A Simple Way to Prevent Neural Networks from Overfitting}.
\newblock {\em Journal of Machine Learning Research}, 15:1929--1958, 2014.

\bibitem{Tagasovska19}
N. Tagasovska and D. Lopez-Paz.
\newblock {Single-Model Uncertainties for Deep Learning}.
\newblock In {\em Advances in Neural Information Processing Systems}, 2019.

\bibitem{Tsang05}
I.~W. Tsang, J.~T. Kwok, P.M. Cheung, and N. Cristianini.
\newblock {Core Vector Machines: Fast SVM Training on Very Large Data Sets}.
\newblock {\em Journal of Machine Learning Research}, 2005.

\bibitem{Vandenhende21}
S. Vandenhende, S. Georgoulis, W.~Van Gansbeke, M. Proesmans, D. Dai, and
  L.~Van Gool.
\newblock {Multi-Task Learning for Dense Prediction Tasks: A Survey}.
\newblock {\em IEEE Transactions on Pattern Analysis and Machine Intelligence},
  2021.

\bibitem{Vandenhende20}
S. Vandenhende, S. Georgoulis, and L.~V. Gool.
\newblock {Mti-Net: Multi-Scale Task Interaction Networks for Multi-Task
  Learning}.
\newblock In {\em European Conference on Computer Vision}, 2020.

\bibitem{Wang20i}
J. Wang, K. Sun, T. Cheng, B. Jiang, C. Deng, Y. Zhao, D. Liu, Y. Mu, M. Tan,
  X. Wang, et~al.
\newblock {Deep high-resolution representation learning for visual
  recognition}.
\newblock {\em IEEE Transactions on Pattern Analysis and Machine Intelligence},
  2020.

\bibitem{Wen2020a}
A. Weller and T. Jebara.
\newblock {Approximating the Bethe Partition Function}.
\newblock In {\em Uncertainty in Artificial Intelligence}, 2014.

\bibitem{Xu18c}
D. Xu, W. Ouyang, X. Wang, and N. Sebe.
\newblock {Pad-Net: Multi-Task Guided Prediction-And-Distillation Network for
  Simultaneous Depth Estimation and Scene Parsing}.
\newblock In {\em Conference on Computer Vision and Pattern Recognition}, pages
  675--684, 2018.

\bibitem{Yang16d}
Y. Yang and T.M. Hospedales.
\newblock {Trace Norm Regularised Deep Multi-Task Learning}.
\newblock In {\em arXiv Preprint}, 2016.

\bibitem{Yoo19}
D. Yoo and I.S. Kweon.
\newblock {Learning Loss for Active Learning}.
\newblock In {\em Conference on Computer Vision and Pattern Recognition}, 2019.

\bibitem{Zhang19f}
Z. Zhang, Z. Cui, C. Xu, Y. Yan, N. Sebe, and J. Yang.
\newblock {Pattern-Affinitive Propagation Across Depth, Surface Normal and
  Semantic Segmentation}.
\newblock In {\em Conference on Computer Vision and Pattern Recognition}, 2019.

\end{thebibliography}
}

\clearpage
\appendix

\section{Supplementary Material}

In this appendix, we describe in more detail how we compute uncertainties for all modalities and justify using the same min-max scheme for all types of Entropy. We then justify computing the normalization constants once and then fixing them throughout the rest of the learning process. Finally, we provide additional training details and discuss the baselines in more detail.

\subsection{Entropy Computation for Different Modalities}
\label{app:unc}

As discussed in Section~\ref{sec:choosing}, to compute uncertainties, we train our model using MC-Dropout~\cite{Gal16a} and perform $D$ forward passes with dropout activated. From the means and variances of these predictions, we  first estimate \textit{pixel-wise} uncertainties and then final $u_{i}^{k}$ \textit{image-wise} uncertainties. We discuss both stages of the process below.

\subsubsection{Pixel-wise Uncertainty Estimation.} 
\label{app:unc-pix}

To estimate the uncertainty for classification tasks such as segmentation, saliency estimation, edge detection, we compute the softmax of $\{p_{d}(y|x)\}_{d=1}^{D}$ and then Shannon's entropy~\cite{Shannon48} of the mean prediction for each pixel. We write
\begin{align}
	\mathbb{H}_{S}[y_i|x]& := - \sum_c \widetilde{p}(y_i=c|x) \log{\widetilde{p}(y_i=c|x)} \; , \\
	&\text{ where } \; \widetilde{p}(y_i|x) = \frac{1}{D} \sum_{d=1}^D p_{d}(y_i|x) \; , \nonumber
	\label{eq:pixel-uncertainty}
\end{align}
%
%
$c$ denotes the possible classes, and $p_{d}(y_i|x)$ is the softmax output for class $c$ on input $x$ at the $i$th pixel.

For regression tasks, such as depth or normals estimation, we use the predictive variance computed from MC-Dropout and calculate the differential entropy following Gaussian assumption for the prediction. For a multivariate Gaussian distribution, this is
\begin{equation}
    \mathbb{H}_{D}[y_i|x] := \frac{N}{2} \log(2\pi) + \frac{1}{2} \log \det{\Sigma} + \frac{1}{2} N \; ,
\label{eq: diff}
\end{equation}
where $N$ is the dimensionality of the output ($N=1$ for depth and $N=3$ for normals) and $\Sigma$ is the covariance matrix estimated from MC-Dropout samples.
 
\subsubsection{Image-wise Uncertainty Estimation.} 
\label{app:unc-img}

For each task and each image, we compute the mean entropy of all pixels so that we end up with a single value for each image / modality pair. Below we provide the intuition behind this choice. 

For simplicity, let us a consider the regression case on a depth estimation example. In this particular case, prediction is $y \in \mathbb{R}^{H \times W}$, where $H$ and $W$ are image height and width, respectively. For analysis purposes, we unfold the depth map $y$ into a 1D vector with $d = H \cdot W$ elements and assume that the predictive distribution is Gaussian: $p_{\theta}(y|x) \sim \mathcal{N}(\mu \in \mathbb{R}^{d}, \Sigma \in \mathbb{R}^{d \times d})$. In major vision applications $d > 1000$ and a robust estimation of $\Sigma$ is not possible, therefore, we also assume $\Sigma$ to be diagonal. Given Eq.~\ref{eq: diff}, the final entropy for $p_{\theta}(y|x)$ becomes
\begin{equation}
    \mathbb{H}_{D}[y_i|x] \propto \log{\det{\Sigma}} = \log{\prod_{i=1}^{d} \sigma_{i}} = \sum{\log{\sigma_{i}}} \;
\label{eq: approx sigma}
\end{equation}
where $\sigma_{i}$ is the variance predicted for the $i$th pixel. Eq.~\ref{eq: approx sigma} demonstrates that $\mathbb{H}_{D}[y_i|x]$ is proportional to the sum of $\log{\sigma_{i}}$ and therefore could be approximated by the average of the pixels' entropies.

\subsection{Shannon's vs Differential Entropies}

Though being similar in form, Shannon's and differential entropies are very different and cannot be compared directly. The min-max scheme of Section~\ref{sec:choosing} has proved to be effective to align Shannon's entropies~\cite{Kumar86}. We now show that this also holds for differential ones and that, after normalization, both entropy types are aligned.

\begin{figure}[h!]
    \includegraphics[width=0.49\textwidth]{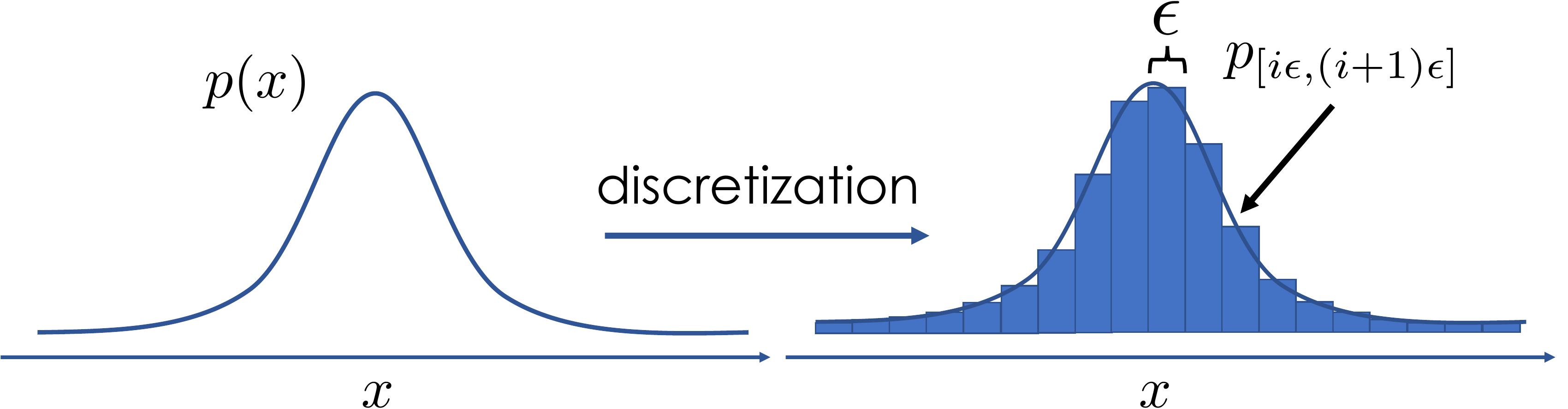}
    \centering
    \caption{\textbf{Discretization of a continuous distribution.} Taking the original continuous distribution $p(x)$, we split it into bins of size $\epsilon$ and assign probability $p_{[i\epsilon, (i+1)\epsilon]}$ to each of those bins.
    }
    \label{fig: discretization}
\end{figure}
Formally, we start with a continuous distribution $p(x)$ and discretize it using a discretization step $\epsilon$, as shown in  Fig.~\ref{fig: discretization}. The discretized distribution is categorical and we can calculate its conventional Shannon's entropy
\begin{align}
H = &\sum_i p_{[i\epsilon, (i+1)\epsilon]}\log\left(\tfrac{1}{p_{[i\epsilon, (i+1)\epsilon]}}\right)\nonumber\\
\text{where } \; & p_{[i\epsilon, (i+1)\epsilon]} = \int_{i\epsilon}^{(i+1)\epsilon} p(x) dx \; .
\end{align}
In the limit $\epsilon \to 0$, the discretized distribution converges to the original one $p(x)$. Hence, using the approximation $p_{[i\epsilon, (i+1)\epsilon]} \approx p(i\epsilon)$ and applying the limit to the entropy $H$ yields 
\begin{align*}
\lim_{\epsilon\to0} \sum_i & p_{[i\epsilon, (i+1)\epsilon]}
\log\left(\tfrac{1}{p_{[i\epsilon, (i+1)\epsilon]}}\right)
= \\
= &\lim_{\epsilon\to0}
\sum_{i} p(i \epsilon)\epsilon
\log\left(\tfrac{1}{p(i \epsilon)\epsilon}\right) = \\
= &\lim_{\epsilon\to0}
\left(\sum_{i} p(i \epsilon)\epsilon
\log\left(\tfrac{1}{p(i \epsilon)}\right)
+ \log(1/\epsilon) \right) = \\
= &
\int_x
p(x)
\log\left(\tfrac{1}{p(x)}\right) dx
+ \lim_{\epsilon\to0}\log(1/\epsilon) \: .
\label{eq: entropies}
\end{align*}
As a result, given a small $\epsilon$ we can rewrite this approximation as 
\begin{small}
\begin{equation}
    -\int_x p(x) \log{p(x)} dx
    \approx
    -\sum_i p_{[i\epsilon, (i+1)\epsilon]}
    \log{p_{[i\epsilon, (i+1)\epsilon]}} + C \; ,
\label{eq: approx}
\end{equation}
\end{small}
where $C$ is an integration constant. In other words, Eq.~\ref{eq: approx} makes it possible to approximate the differential entropy (left side) with the Shannon entropy (right side) up to a constant. Hence,  the min-max scheme is warranted.

\subsection{Fixed Normalization Parameters}

As mentioned in Section~\ref{sec:choosing}, we compute the normalization parameters $u_{\rm{max}}^{k}$ and $u_{\rm{min}}^{k}$ once during the first active learning iteration and then reuse them for all subsequent ones. We now justify this design choice. 

First, let us consider a hypothetical setup in which normalization parameters $u_{\rm{max}}$ and $u_{\rm{min}}$ {\it are} updated at each active learning step. Normalized entropies will be scaled to the same interval $[0, 1]$ at each iteration. Therefore, even if the model is much more uncertain about one modality $K_{1}$ than about the other $K_{2}$, they will be mapped to the same scale. \pale{} tries to find the most uncertainty image / modality pairs and the above-mentioned behavior would make this search impossible. 

Second, let us consider our normalization scheme from a Bayesian perspective. Every time we add new samples to $\bL$, the uncertainty on unlabelled samples $\bU$ should decrease. This stems from the fact that $\mathbb{H}(Y) \ge \mathbb{H}(Y|X)$, which means that conditioning can only decrease the entropy. In other words, 
\begin{align}
\mathbb{H}(&P(y^{k}|x, \bL)) > \mathbb{H}(P(y^{k}|x, \bL + (y_{new}, x_{new})))\nonumber \; ,\\ 
&\text{ where } \; x\in \bU \; \text{ and } \; u^k = \mathbb{H}(P(y^{k}|x, \bL)).
\end{align}
Our final entropy values $\widetilde{u}^{k}$ follow this behavior, making comparison of uncertainties robust and well-defined. In contrast, normalization schemes with changing parameters lose these guaranties.  

\subsection{Training Details}

We now provide details about training and architectural choices for NYUv2 and PASCAL experiments. For both datasets we use PAD-Network~\cite{Xu18c} with HRNet W18~\cite{Wang20i} backbone. Initial training set for NYUv2 consists of 100 images with all of the target labelled, while initial training set for PASCAL consists of 1000 fully labelled images.
For both datasets, during each active learning iteration we run 100 training epochs using popular Adam~\cite{Kingma14a} optimizer with $10^{-4}$ learning rate and $10^{-4}$ weight decay. Learning rate is being updated with polynomial scheduler as 
$$\text{learning\_rate} = \text{learning\_rate} * (1 - x / 100)^{0.9},$$
where $x$ is the number of current epoch. For both datasets, we use weighted sum of task losses as $\mathcal{L}_{\text{total}} = \sum^{N}_{i=1} w_{i} \mathcal{L}_{i} $. For NYUv2 these weights are $1.0$ for semantic segmentation, $1.0$ for depth estimation and $10.0$ for normals estimation. For PASCAL these are $1.0$ for semantic segmentation, $2.0$ for human parts segmentation, $5.0$ for saliency, $50.0$ for edge detection and $10.0$ for normals estimation.

\subsection{Active Learning Baselines}

As mentioned in the experiment section, we compare against several popular active learning baselines to validate our method.
For \textit{Random} baseline, we pick 80 new samples for NYUv2 and 300 for Pascal arbitrary without considering any uncertainty or confidence values. For \textit{Coreset} approach, we employ K-center greedy solver from~\cite{Sener17} and choose the same number of samples as for \textit{Random}.
For \textit{Learning Loss} baseline, we generate deep features from network's encoder and use them to predict loss of the model with small-scale feed-forward network which is trained alongside the original model. For \textit{RBAL}, we replicate all of the steps from~\cite{Reichart08} and generate uncertainties for each of the tasks and utilize ranking-based procedure to find the most complicated samples. For \pale{}, we generate uncertainties as it was described in Section~\ref{app:unc}, apply normalization procedure and pick the most complicated sample / modality pairs -- 240 for NYUv2 and 1500 for PASCAL datasets.

\end{document}